\documentclass[final,3p,times,twocolumn]{elsarticle}
\usepackage{amssymb}
\usepackage{float}
\usepackage{subfig}
\usepackage{tikz}
\usepackage{multirow}
\usepackage{booktabs}
\usepackage{pifont}
\makeatletter
\g@addto@macro{\endtabular}{\rowfont{}}% Clear row font
\makeatother
\newcommand{\rowfonttype}{}% Current row font
\newcommand{\rowfont}[1]{% Set current row font
\gdef\rowfonttype{#1}#1\ignorespaces%
}
\makeatother
\usepackage[printonlyused,withpage]{acronym}

\newcommand{\cmark}{\ding{51}}%

\def\arrvline{\hfil\kern\arraycolsep\vline\kern-\arraycolsep\hfilneg}
\newcommand*\fullcirc[1][1ex]{\tikz\fill (0,0) circle (#1);} 
\newcommand*\emptycirc[1][1ex]{\tikz\draw (0,0) circle (#1);} 
\usepackage{amsmath}
\newcommand{\ubold}[1]{\fontseries{b}\selectfont #1} % non-extended bold

\journal{Medical Image Analysis}

\begin{document}

\begin{frontmatter}



\title{Multicentric thrombus segmentation using an attention-based recurrent network with gradual modality dropout} %% Article title


\author[label1]{Sofia Vargas-Ibarra} %% Author name
\author[label1]{Vincent Vigneron} %% Author name
\author[label1]{Hichem Maaref} %% Author name
\author[label2]{Sonia Garcia-Salicetti} %% Author name

\affiliation[label1]{organization={Universite Evry Paris-Saclay, IBISC},%Department and Organization
            addressline={34 Rue du Pelvoux}, 
            city={Evry-Courcouronnes},
            country={France}}
\affiliation[label2]{organization={SAMOVAR, T\'el\'ecom SudParis, Institut Polytechnique de Paris},%Department and Organization
            addressline={34 Rue du Pelvoux}, 
            city={Palaiseau},
            country={France}}

%% Abstract
\begin{abstract}
Detecting and delineating tiny targets in 3D brain scans is a central yet under-addressed challenge in medical imaging. In ischemic stroke, for instance, the culprit thrombus is small, low-contrast, and variably expressed across modalities (e.g., susceptibility-weighted/T2* blooming, diffusion restriction on DWI/ADC), while real-world multi-center data introduce domain shifts, anisotropy, and frequent missing sequences. We %recast thrombus segmentation as a generic {small-object detection in 3D neuroimaging} problem and
introduce a methodology that couples an  \ac{upatt}, a training schedule that progressively increases the difficulty of hetero-modal learning, with gradual modality dropout, \ac{upatt} aggregates context across slices via recurrent units (2.5D) and uses attention gates to fuse complementary cues across available sequences, making it robust to anisotropy and class imbalance. Gradual modality dropout systematically simulates site heterogeneity, noise, and missing modalities during training, acting as both augmentation and regularization to improve multi-center generalization. On a monocentric cohort, our approach detects thrombi in $>$90\% of cases with a Dice score of 0.65. In a multi-center setting with missing modalities, it achieves $\sim$80\% detection with a Dice score around 0.35. Beyond stroke, the proposed methodology directly transfers to other small-lesion tasks in 3D medical imaging where targets are scarce, subtle, and modality-dependent.%
% Text of abstract
\end{abstract}

%\item Multicentric thrombus segmentation in stroke patients using MRI
%\item Training technique to improve the robustness when some modalities are missing
%\item Attention-based architecture to segment the thrombus using the lesion information

%% Keywords

\end{frontmatter}



\acrodef{clstm}[CLSTM]{convolutional long-short-term memory}
\acrodef{mri}[MRI]{Magnetic Resonance Image}
\acrodef{ai}[AI]{artificial intelligence}
\acrodef{ml}[ML]{machine learning}
\acrodef{ann}[ANN]{artificial neural network}
\acrodef{dl}[DL]{deep learning}
\acrodef{mlp}[MLP]{multi-layer perceptron}
\acrodef{mtl}[MTL]{multi-task learning}
\acrodef{stl}[STL]{Single-task learning}
\acrodef{moe}[MOE]{mixture of experts}
\acrodef{nlp}[NLP]{Natural Language Processing}
\acrodef{gmd}[GMD]{gradual modality dropout}
\acrodef{upatt}[UpAttLLSTM]{attention-based recurrent segmentation network}
\acrodef{cv}[CV]{computer vision}
\acrodef{gru}[GRU]{gated recurrent unit}
\acrodef{tl}[TL]{transfer learning}
\acrodef{cd}[CD]{cardiovascular dataset}
\acrodef{cp}[CP]{cerebral palsy}
\acrodef{cga}[CGA]{clinical gait analysis}
\acrodef{lstm}[LSTM]{long-short-term memory}
\acrodef{cnn}[CNN]{convolutional neural network}
\acrodef{mae}[MAE]{mean absolute error}
\acrodef{mse}[MSE]{mean square error}
\acrodef{rmse}[RMSE]{root mean square error}
\acrodef{mse}[MSE]{mean square error}
\acrodef{nrmse}[NRMSE]{normalized root mean square error}
\acrodef{se}[SE]{Standard error}
\acrodef{sd}[SD]{Standard deviation}
\acrodef{tbi}[TBI]{traumatic brain injury}
\acrodef{sci}[SCI]{spinal cord injury}
\acrodef{cp}[CP]{cerebral palsy}
\acrodef{ms}[MS]{multiple sclerosis}
\acrodef{cma}[CMA]{clinical movement analysis}
\acrodef{qga}[QGA]{quantified gait analysis}
\acrodef{mri}[MRI]{magnetic resonance imaging}
\acrodef{ct}[CT]{computed tomography}
\acrodef{chsf}[CHSF]{{Centre Hospitalier Sud-Francilien}}%{Centre Hospitalier Sud-Francilien}
\acrodef{dwi}[DWI]{diffusion-weighted imaging}
\acrodef{nn}[NN]{neural network}
\acrodef{lstm}[LSTM]{long-short time memory} 
\acrodef{llstm}[LLSTM]{Logic long-short time memory}
\acrodef{ais}[AIS]{acute ischemic stroke}
\acrodef{mri}[MRI]{magnetic resonance imaging}
\acrodef{clstm}[CLSTM]{convolutional long-short term memory}
\acrodef{adc}[ADC]{Apparent diffusion coefficient}

\acrodef{ct}[CT]{computed tomography}
\acrodef{chsf}[ ]{{\sf ANONYMOUS FOR REVIEW}}%{Centre Hospitalier Sud-Francilien}
\acrodef{dwi}[DWI]{diffusion-weighted imaging}
\acrodef{nn}[NN]{neural network}
\acrodef{lstm}[LSTM]{long-short time memory} 
\acrodef{llstm}[logic LSTM]{logic long-short time memory}
\acrodef{flair}[FLAIR]{fluid-attenuated inversion recovery}
\acrodef{ct}[CT]{computerized tomography}
\acrodef{hosan}[hos X]{hospital X}
\acrodef{hos}[Centre Hospitalier Sud Francilien]{Centre Hospitalier Sud Francilien}
\acrodef{chsfan}[Dataset 1]{Proximal}
\acrodef{mataran}[Dataset 2]{Distal}
\acrodef{chsf}[CHSF]{Proximal}
\acrodef{matar}[MATAR]{Distal}
\acrodef{main}[MAIN]{masked adaptive instance normalization}
\acrodef{cnn}[CNN]{convolutional neural network}
\acrodef{dnn}[DNN]{deep neural network}
\acrodef{gan}[GAN]{generative adverserial network}
\acrodef{jhu}[JHU]{Johns Hopkin's university}
\acrodef{b0}[B0]{magnetic field strength}
\acrodef{flair}[FLAIR]{fluid-attenuated inversion recovery}

\acrodef{swi}[SWI]{susceptibility-weighted imaging}

\acrodef{swan}[SWAN]{susceptibility weighted angiography }
\acrodef{pwi}[PWI]{Pulse wave imaging}
\acrodef{tof}[TOF]{Time-of-flight}
\acrodef{asl}[ASL]{Arterial Spin Labelling}

\section{Introduction}
Stroke is responsible of more than 22 percent of deaths globally and is the first cause of disability in adults.  When the cause of the stroke is the internal bleeding in the brain, the stroke is denoted as hemorrhagic. Otherwise, when a blood clot (thrombus) blocks an artery, the stroke is named ischemic stroke. The last one represents 62\% of cases and restricting the blood circulation, it initiates brain tissue damage. The damaged area, denoted as lesion, consists on a recoverable area (penumbra) and on death tissue (core).   In an emergency case, the pathology is in the hyper-acute phase ($\leq$ 6h from the onset of the symptoms) which is critical for saving lives and improving
life expectancy of patients.  The thrombus location, being the cause of the cerebral attack, is a crucial information for choosing the treatment. Nowadays, all the stroke characterization is manually done  by the neurologists using brain scans, either Computerized Tomography (CT)  or \ac{mri}. 

We propose  a full automatic method based on deep learning for the thrombus segmentation in a multicentric scenario. We define a recurrent network based on an attention operation (UpAttLLSTM) between MRIs modalities for the thrombus segmentation. Thanks to our proposed method "gradual modality dropout" the architecture  produces a robust performance when there are variations on the modalities or missing ones due to the brain scan machine. 
\label{sec1}
%% Labels are used to cross-reference an item using \ref command.
\section{Related works}
 Due to the limited public datasets for the thrombus segmentation, some approaches use classical image processing methods.  In~\cite{thrombict}, an automatic thrombus segmentation method for CT scans uses the contralateral vasculature segmentation for detecting the occlusion and  a region-growing method based on the intensity values produces the thrombus mask. As the mentioned method  requires a manual annotation,  in~\cite{thrombict2}, some possible objects are first automatically extracted using a threshold over the Hounsfield Units from CT scans. Their classification is obtained by a Random Forest classifier using intensity-based, geometrical-based, and location-based features.

 Deep learning models aim to obtain a more robust prediction being independent from the hyperparameter choices. Polar-UNet~\cite{polarunet} uses baseline Unet to segment thrombi in CT scans, restricting the volume of interest to the brainstem.  In~\cite{hemisphere}, the symmetric property of a healthy brain is exploited. Using symmetric patches from the two brain hemispheres, a CNN classifier predicts if they are symmetric and if there is a hyperdense artery sign (HAS). If HAS is found to be asymmetric, the patches are considered as thrombus and the segmentation is obtained through another network. In addition, in~\cite{thrombict3}, a two-stage method is proposed based on Unet. It uses deep supervision and multiscale training for segmenting the thrombi in CT scans. Using MRIs, a Multidirectional Unet is exploited in~\cite{jon}: one architecture is defined per axe - axial, coronal, and sagittal planes- and the final segmentation is obtained by merging the three resulting masks. Finally, in~\cite{thesis} a recurrent method denoted as Logic LSTM is proposed, which uses the sagittal plane recurrently to obtain the thrombi. Each slice prediction is obtained from the previous and following slices, exploiting the dense property of the thrombi and mimicking the doctor's procedure. 

Most of these methods are focused on monocentric datasets  and so the multicentric robustness is not evaluated.  Apart from the distribution differences between centers, for the thrombus segmentation, different brands produce different modalities, arising a missing data problem. To manage absent data, some encoding-decoding approaches have been proposed, where the decoding reconstructs the input modalities and the latent space produces the segmentation. Unified representation network~\cite{unified} learns to predict with a variable number of input modalities.  By computing the mean of the encoded representations, the model  can produce an  output that is independent from the number of available modalities. In~\cite{missinglatent}, a Latent Correlation Representation is proposed, where a linear combination of the encoded features is computed and then the missing modality can be described with the available ones.  Generative Adversarial Network (GAN) \cite{gans} allows to generate the missing data. In~\cite{adversial}, the generator produces the synthesised versions of the missing modalities, which are unable to be distinguished between fake or real by the discriminator, allowing to train a model under a non-missing scenario. In addition, some methods use a co-training strategy, where two architectures are trained in parallel: one with all modalities and another with some absent ones. MMCFormer~\cite{mmcformer} trains these two networks in parallel using some distillation modules and SMU-Net~\cite{stylematch} uses a style and content-matching mechanism to improve the segmentation in a missing data case. The style-matching mechanism maintains the texture information, whereas the content-matching module aims to maintain the structural and semantic characteristics. 

Finally, Modality dropout~\cite{modalitydropout} turns off a modality during training, improving the model robustness under missing data scenarios. In~\cite{dropmultidata}, authors compare  several architectures performances. The best performances are obtained with MultiUnet, which is a Unet-based model where during training some modalities have been dropped. An extension of this method is ModDrop++~\cite{moddropplus}, where a dynamic head is included before the first convolutional layer.  With a modality code vector, the kernel weights of the first convolutional layer are scaled. They are multiplied by the corresponding scale factor and  allows the network to be adapted depending on the initial conditions: the dynamic filter is different depending first on whether a modality is missing and in which position it is and  an intra-subject co-training is included.

All the proposed methods help to improve the missing data without compromising the full modality scenario. Some of them are model-agnostic, and others include specific training settings. However, netiher of them have been tested in a realistic scenario, for example, in a multicentric scenario where the different brain scan machines produce  different modalities, some of them being absent.  In this work we propose a specific segmentation architecture for the thrombus that is robust in a realistic multicentric scenario, where some modalities are missing.

\section{Datasets}
Data from two multi-center public datasets ISLES~\cite{isles2022}, JHU~\cite{jhu},  three single-center ones coming from Centre Hospitalier Sud Francilien~\cite{chsf} (CHSF and MATAR/MATAR2) and a private multicentric dataset coming from hospitals in the Parisian region (FOCH) are present in this study. They consist on brain MRIs of patients suffering from an ischemic stroke. 

Each dataset introduces unique features and challenges.  ISLES, JHU and FOCH exhibit significant variability in imaging devices, protocols, and patient demographics, providing an opportunity to train robust and generalizable algorithms. On the other hand,  the monocentric datasets offer more controlled environments.  The datasets differ in their characteristics:  quantity of data, type of stroke, location, volume, and image resolution. Table \ref{table:complete_dataset} summarises the dataset characteristics. 

\setlength{\tabcolsep}{2.5pt}

\begin{table*}[!htb]
%\floatconts
\centering

{\scalebox{0.8}{\begin{tabular}{lcccccc}
\toprule
& CHSF& MATAR & MATAR2 &ISLES & JHU &FOCH \\ 
\midrule

Annotated pat. & $65$& $125$& $156$ &$250$& $2888$ &$43$ \\ 
   % Age (years)  & 74.33  $\pm$ 0.74&72.95 	$\pm$ 1.6 \\

 % Female/male  & 57.7\%/43.3$\%$ & 54.5\%/46.5$\%$   \\
  %  Hypertension  & 62$\%$ & 59$\%$\\
  %  Current smokers  &11$\%$  & 13$\%$ \\
    %Initial NIHSS  &  4.9$\pm$0.51 &\\
   % DWI-ASPECTS\footnotemark  & 3.69 (0-10) & 9 (7.75-10)\\
%Diffusion mod. %&DWI,ADC,B0&DWI,ADC,B0&DWI,ADC,B0&DWI,ADC&DWI,ADC,B0&DWI,ADC,B0 \\ 
%Susceptibility mod. &SWAN,PHASE&SWAN,PHASE&SWAN,PHASE&-&SWI, MagSWI&SWAN (28\%), SWI (72\%)\\ 
%Other mod. & FLAIR, TOF &FLAIR, TOF &FLAIR, TOF &FLAIR &FLAIR, TOF &FLAIR, TOF \\

Multicentric & & && \cmark& \cmark &\cmark\\
Multiequipement & & && \cmark& \cmark &\cmark\\

Ischemic &$100\%$&$100\%$&$100\%$&$100\%$&$65\%$&$100\%$\\
Hemorrhagic &$0\%$&$0\%$&$0\%$&$0\%$&$35\%$&$0\%$\\

Control \ac{mri} & & \cmark&\cmark & & &\\
Only proximal & \cmark&& & & & \\
Only distal & & \cmark &\cmark& & & \\
Lesion volume (mL) & $31.77$ & $5.01$ &$5.97$& $23.38$& $23.15$ & -\\
Thrombi volume (mL) & $0.23$ & $0.08$&$0.09$ &- & - & $0.31$\\
Diffusion mod. & ADC, B0, DWI &ADC, B0, DWI & ADC, B0, DWI & ADC, DWI& ADC, B0, DWI &  B0, DWI \\
Susceptibility mod. & SWAN,PHASE & SWAN,PHASE & SWAN,PHASE & - &- & SWI(28\%), SWAN (72\%) \\
\multirow{3}{*}{\ac{mri} machines} & \texttrademark{GE} & \texttrademark{GE}& \texttrademark{GE} &\texttrademark{Philips}, \texttrademark{Siemens},  & \texttrademark{Siemens},   & \texttrademark{Philips},  \texttrademark{GE},  \\ 
& & & &  \texttrademark{Philips},  \texttrademark{GE} & \texttrademark{Toshiba}&\texttrademark{Siemens} \\
\hline 
\end{tabular}}}
\caption{Dataset description. The study incorporates four distinct datasets, each with unique characteristics. The multi-centric nature of the public datasets (JHU, ISLES, FOCH) ensures diversity in patient populations and imaging protocols. In contrast, the specialised private datasets offer targeted insights into specific types of ischemic strokes. The two private ones consist only of hyper-acute ischemic strokes, one with distal occlusions and the other with proximal ones. \label{table:complete_dataset}}
\end{table*}
The monocentric datasets come from  1.5 T and 3 T \ac{mri} scanners from \texttrademark{GE} Healthcare. In FOCH, the scans come from several machines:  \texttrademark{Philips} represents 16\% of the images, \texttrademark{GE} the 44\% and \texttrademark{Siemens} the 41\%. ISLES is captured on diverse \ac{mri} scanners such as 3T \texttrademark{Philips} (Achieva, Ingenia), 3T \texttrademark{Siemens} (Verio), and 1.5T \texttrademark{Siemens MAGNETOM} (Avanto, Aera) and JHU includes scanners from  1.5 T (61\%) and 3 T (39\%) machines from \texttrademark{Siemens} (91.66\%), \texttrademark{Toshiba}  (0.21\%), \texttrademark{Philips}  (7.20\%), and \texttrademark{GE} (0.93\%).    Concerning the delay between the beginning of the ischemic stroke symptoms and MRI, just CHSF, MATAR, MATAR2, and FOCH consist of only hyper-acute strokes, with a delay time of less than 4.5 hours. ISLES has acute and subacute strokes (until 3 weeks after a stroke), and JHU images were acquired between 6 and 24 hours post-symptom onset, being in the acute phase. In addition, only distal occlusions are available in MATAR and MATAR2. 

The total number of annotated patients differs a lot between the different datasets: 2888 in JHU (where only 65\% of them are ischemic), 250 in ISLES, 65 in CHSF, 125 in MATAR, 156 in MATAR2, and 43 in FOCH. In FOCH the full dataset has 87 patients and MATAR2, 281. The lesion segmentation is not available for FOCH and the thrombus is only segmented in the monocentric and FOCH dataset.  The lesion volume in CHSF dataset is of 31.77 ml on average, 5.01 ml for MATAR of 5.01 ml, 6ml for MATAR2 and 23ml for  ISLES and JHU datasets. The thrombus volume  for CHSF is 0.23ml on average,  0.08ml for MATAR,0.09ml for MATAR2 and 0.31ml for FOCH.

Being focused on both segmentations, only two types of modalities are used in this study: diffusion modalities (\ac{dwi}, \ac{adc} and \ac{b0}) and susceptibility ones (\ac{swan} and  PHASE). The diffusion ones  are available in all datasets except for ISLES where B0 is missing and for FOCH, where \ac{adc} is missing and the susceptibility ones  are  only available in the monocentric datasets and \ac{swi} or \ac{swan} in FOCH. In the diffusion ones the lesion is visible and in the susceptibility ones the thrombi. 

Finally, due to the different machines' parameters, image resolutions are different in each dataset.  For the diffusion modalities, the sizes in ISLES are in the ranges of (112-256)$\times$(112-256)$\times$(30-74) voxels. For JHU, (128-384)$\times$ (128-384)$\times$(21-81) voxels and of (128-512)$\times$(128-512)$\times$(20-52) for FOCH.  Finally, in the three monocentric datasets, the diffusion modalities are of sizes  256$\times$256$\times$(24-40). The susceptibility modalities' resolution also differs in the following ranges (138-512)$\times$(192-512)$\times$(28-144) for JHU, (192-1024)$\times$(256-1024)$\times$(24-130) for FOCH and 512$\times$512$\times$(72-232) for CHSF, MATAR, and MATAR2.

%\begin{table}[!htb]
%\floatconts
%\centering

%{\scalebox{0.65}{\begin{tabular}{c|ccccccccccc}
%\toprule
%Dataset& \ac{dwi}&  \ac{adc} &  \ac{b0} & \ac{flair} &  \ac{tof}&\ac{swan}&\ac{swi}&PHASE  %&SWIMag&SWIPha &T2w\\ 
%\midrule

%CHSF, MATAR, MATAR2 &\cmark &\cmark &\cmark& \cmark& \cmark &\cmark & &\cmark\\
%ISLES&\cmark &\cmark &\\
%JHU &\cmark &\cmark &\cmark& \cmark&  && 66.6\% & & 56.8\%& 56.8\%    & \cmark\\
%FOCH &\cmark & &\cmark& \cmark& \cmark &28\% & 72\%& \\
%\hline 
%\end{tabular}}}
%\caption{Modality availability per dataset \label{tab:modalitiesperdata}}
%\end{table}

\section{Methodology}
\label{sec:gradualmod}
\begin{figure}[!htb]
\centering
\begin{minipage}{0.15\textwidth}

    \centering
    \includegraphics[width=1.0\textwidth]{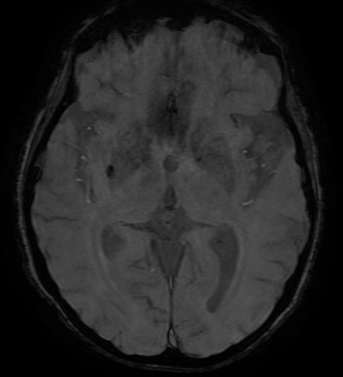}
    \caption{SWAN}
    \label{fig:swan1}

\end{minipage}
\begin{minipage}{0.15\textwidth}

    \centering
    \includegraphics[width=1.0\textwidth]{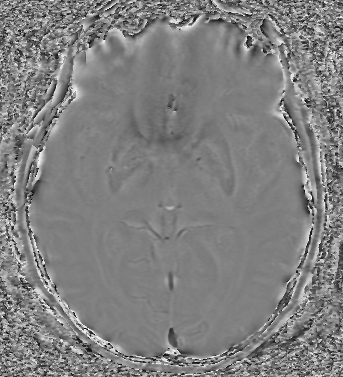}
    \caption{PHASE}
    \label{fig:swan2}

\end{minipage}

\begin{minipage}[t]{0.15\textwidth}

    \centering
    \includegraphics[width=1.0\textwidth]{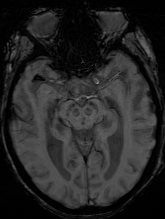}
    \caption{SWI}
    \label{fig:swi1}

\end{minipage}
\begin{minipage}[t]{0.15\textwidth}

    \centering
    \includegraphics[width=1.0\textwidth]{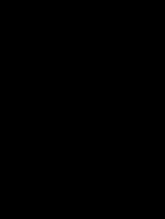}
    \caption{PHASE is absent}
    \label{fig:swi2}

\end{minipage}
\end{figure}

For the thrombus segmentation, the two main MRI brands produce two different modalities:  \texttrademark{Siemens} produces as susceptibility modality SWI while \texttrademark{GE} produces SWAN and PHASE. If a model is trained using SWAN and PHASE and wants to be tested on new centers where only SWI is available, there are two main problems:

\begin{itemize}
\item PHASE modality  is missing (Fig. \ref{fig:swan2} vs Fig. \ref{fig:swi2}).
\item SWI is different than SWAN (Fig. \ref{fig:swan1} vs Fig. \ref{fig:swi1}).
\end{itemize}

\begin{figure*}[!htb]
 \centering
 \includegraphics[width=0.8\textwidth]{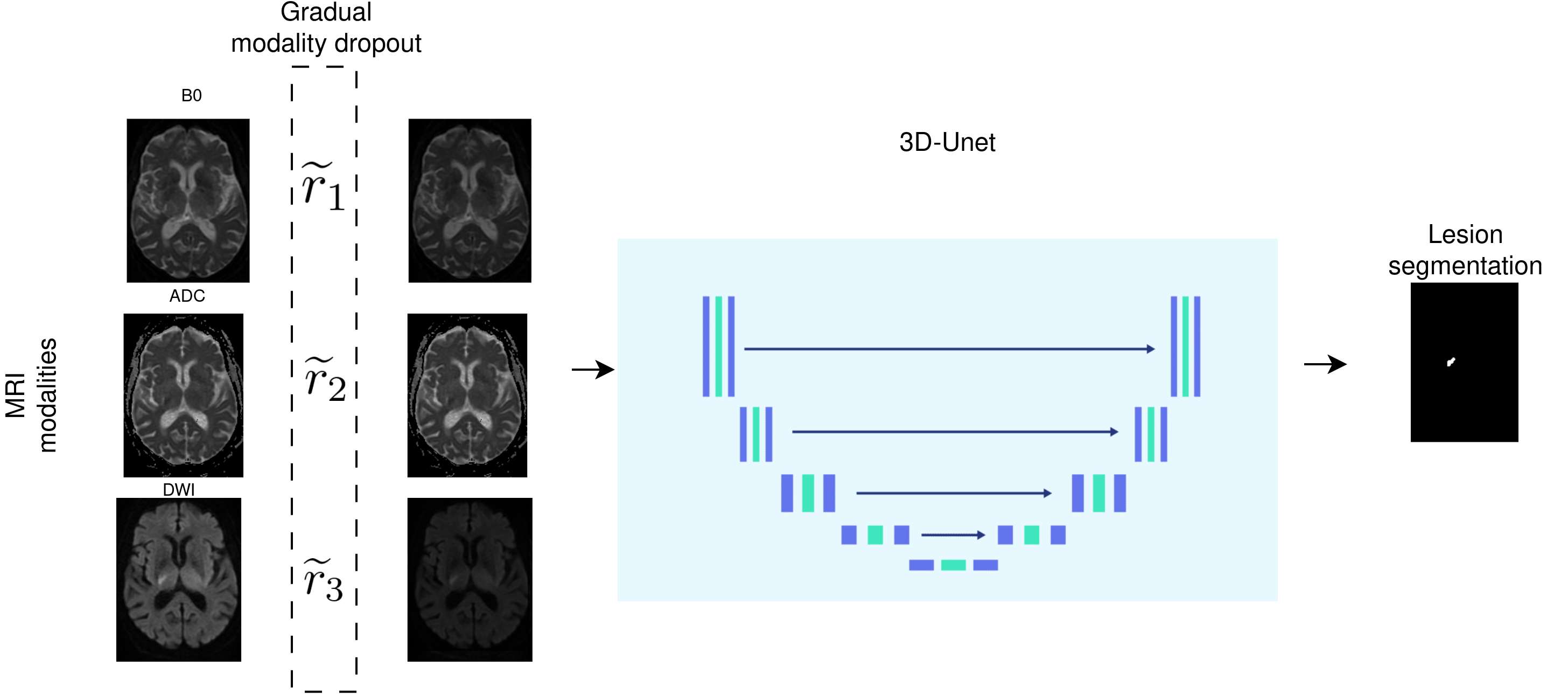}
\caption{Gradual modality dropout. Each modality is multiplied by a value $\widetilde{\mathbf{r}}_{j}$ before entering the segmentation model during the training phase, which regulates the degree of modality dropout. }
\label{fig:Gradual modality dropout}
\end{figure*}

Trying to maximize the input information, we propose an architecture that can segment  the thrombus in both cases: when one modality is absent or when all are available. We propose gradual modality dropout, being model-agnostic, it requires no synthetic modality generation and adds no extra training complexity. Thanks to its regularizer and augmentation aspect, it can produce robust performances when some data is missing. For the thrombus segmentation, we propose a recurrent architecture for the thrombus segmentation, which merges the lesion and thrombus informative modalities using an attention module.  Mimicking the neurologist procedure, once the lesion and thrombus information is merged, the prediction per slice is obtained thanks to the previous and next slices. 

\subsection{Gradual modality dropout}

Dropout~\cite{srivastava2014dropout} is a regularization technique that randomly deactivates certain network connections during the training phase of a \ac{dnn}. For each hidden layer $l$, the feed-forward operation for a hidden unit $i$  is defined as: 
\begin{equation}
\begin{split}
r_{j}^{(l)} \mbox{ } \sim \mbox{ } & \text{Bernouilli}(p) \\
\widetilde{\mathbf y}^{(l)} = \mbox{ } & \mathbf{r}^{l} \mbox{ * } \mathbf{y}^{(l)}\\
\mathbf{z}_{i}^{(l+1)} = \mbox{ } & W_{i}^{l+1}\widetilde{\mathbf{y}}+\mathbf{b}_{i}^{(l+1)}\\
\mathbf{y}_{i}^{(l+1)} =\mbox{ } & f(\mathbf{z}_{i}^{(l+1)})
\end{split}
\end{equation}
where $\mathbf{z}^{(l)}$ is the vector of inputs of layer $l$, $\mathbf{y}^{(l)}$ the vector of outputs from layer $l$, $W^{(l)}$ and $\mathbf{b}^{(l)}$ are the weights and bias at layer $l$,
 $f$ is any activation function, $\mathbf{r}^{(l)}$ is a vector of  Bernoulli random variables, with probability $p$ of being 1, $j \in \{0,\dotsc,n-1\}$, $n$ is the size of the vector $\mathbf{y}^{(l)},l \in \{0,\dotsc, L-1\}$, $L$ is the number of hidden layers and $*$ denotes an element-wise product.

Modality dropout applies dropout specifically to input modalities~\cite{modalitydropout} by zeroing out entire input images with a Bernoulli probability $(1-p)$.
For a segmentation model $F$ that uses input modalities $\mathbf x$, where  $n$ is the number of modalities and the final output $\mathbf y$ becomes: 
\begin{equation}
\begin{split}
 r_{j} \mbox{ } \sim& \mbox{ } \text{Bernouilli}(p) \\
\widetilde{\mathbf x}\mbox{ } =& \mbox{ } \mathbf{r} \odot {\mathbf x}\\
{\mathbf y} \mbox{ } =& \mbox{ } F(\widetilde{{\mathbf x}})
\label{eq:moddrop}
\end{split}
\end{equation}

where $\odot$ is the Hadamard product. $\widetilde{\mathbf x}$ becomes a transformation of the input, where the modalities are zeroed with probability $(1-p)$, as $\mathbf{r}_{j}$ is equal to 0 with that probability. Thanks to that enhanced input, $F$ becomes robust to missing data, meaning black input images. 

To improve training stability, a gradual modality dropout approach is introduced \cite{sofia2}. Here, dropout is incrementally applied overtraining, allowing a smooth transition from fully visible to fully dropped-out inputs. This is mathematically represented as:

\begin{equation}
\begin{split}
r_{j} \mbox{ } \sim \mbox{ } & \text{Bernouilli}(p) \\
\widetilde{ r}_{j} \mbox{ } = \mbox{ } &\begin{cases}
  {1}, & \text{if $r_{j}=1$}\\
   {g}_{j}(t), & \mbox{otherwise}
\end{cases} \\
\widetilde{ \mathbf{x}} \mbox{ } = \mbox{ } & \widetilde{ \mathbf{r}} \odot \mathbf{x}\\
 \mathbf{y} \mbox{ } = \mbox{ } & F(\widetilde{ \mathbf{x}})
\label{eq:gradmod}
\end{split}
\end{equation}
where $ {g}_{j}(t)$ is a time-dependent function that gradually reduces from 1 (no dropout) to 0 (full dropout). This enables the model to transition smoothly from full-information inputs, without changing any contrast in the image, to training with partially dropped-out modalities, where the contrast is smoothly reduced, enhancing its ability to handle missing data.

Modality dropout is applied selectively, affecting only the variables where $r_{j} \neq 1$ (see Fig. \ref{fig:Gradual modality dropout}). Here, $\widetilde{r}$ is an $n$-dimensional vector. When all values of $\widetilde{r}$ are set to one, the input image remains fully intact, and $g_{j}(t) = 1$. During training, $\widetilde{r_{j}}$ gradually decreases toward a zero vector, ultimately blacking out the image and allowing the model to transition smoothly from learning with complete modality information to handling partially dropped-out inputs. When $g_{j}(t) = 1$, modality $j$ is fully retained; when $g_{j}(t) = 0$, it is entirely dropped. A summary of this method is shown in \ref{fig:Gradual modality dropout}, using Unet as a segmentation model.

This progressive dropout technique encourages the model to generalize effectively by gradually reducing reliance on specific modalities without sudden disruptions. For this, $g$ must be a decreasing function reaching zero at some time $t$. The specific schedule we used for $g_{j}(t)$ is as follows:
\begin{equation}
g_{j}(t) = \begin{cases}
0.75, & \text{if } t < 0.25T \\
0.5, & \text{if } t < 0.5T \\
0.25, & \text{if } t < 0.75T \\
0, & \text{otherwise}
\end{cases}
\end{equation}
where $T$ represents the total number of epochs. Initially, $g(t) = 0.75$, decreasing to 0.5 after 25\% of epochs and to 0.25 at the halfway point before dropping fully (with $g(t) = 0$) in the final 25\% of epochs.

\begin{figure}[!htbp]
 \centering
\subfloat[DWI with $\widetilde{\mathbf{r}}_{j}$ = 1 ]{ \includegraphics[width=0.2\textwidth]{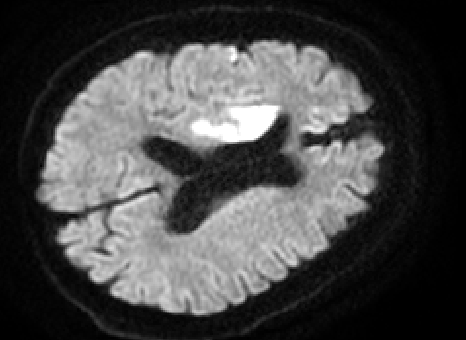}}
\hspace*{0.1in}
\subfloat[DWI with $\widetilde{\mathbf{r}}_{j}$ = 0.75 ]{ \includegraphics[width=0.2\textwidth]{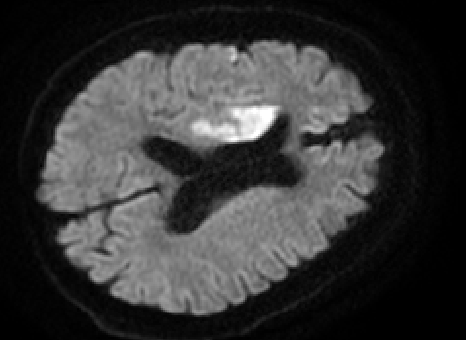}}

 \subfloat[DWI with $\widetilde{\mathbf{r}}_{j}$ = 0.5 ]{ \includegraphics[width=0.2\textwidth]{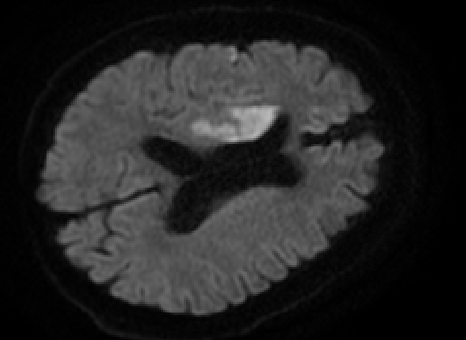}}
 \hspace*{0.1in}
\subfloat[DWI with $\widetilde{\mathbf{r}}_{j}$ = 0.25]{ \includegraphics[width=0.2\textwidth]{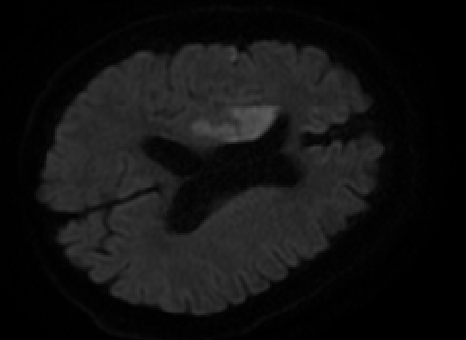}}

\caption{Gradual modality dropout examples using DWI}
\label{fig:gradual}
\end{figure}

Figure \ref{fig:gradual} illustrates how the input image evolves using this dropout function, providing a smoother transition. that enhances the model’s adaptability and robustness. During training, the image contrast is reduced until arriving at a black image, and then mimicking the full dropout technique.

To introduce noise to $ \widetilde{\mathbf{r}} $, random values $\epsilon$ are added from a normal distribution with a mean of 0 and a standard deviation of 0.01:
\begin{equation}
\widetilde{r}_{j} \mbox{ } = \mbox{ } \begin{cases}
1, & \text{if $r_{j} =1$}\\
g_{j}(t) + \epsilon, & \mbox{otherwise}
\end{cases} 
\end{equation}

This approach serves both as data augmentation, where $ \widetilde{\mathbf x} $ becomes an enhanced version of $ \mathbf x $, and as regularisation by allowing the model to adapt to missing modality data gradually. This progressive dropout lets the model predict with less reliance on modality $ j $, potentially learning even when that modality is completely absent. This augmentation occurs only during training; in the inference stage, the model can be tested with either all modalities or some missing (by replacing the absent modality with a black image).

\subsection{UpAttLLSTM for thrombus segmentation}
\begin{figure*}[!ht]

\centering

 \vspace*{-0.1in} 
 \centering
 \includegraphics[width=0.9\textwidth]{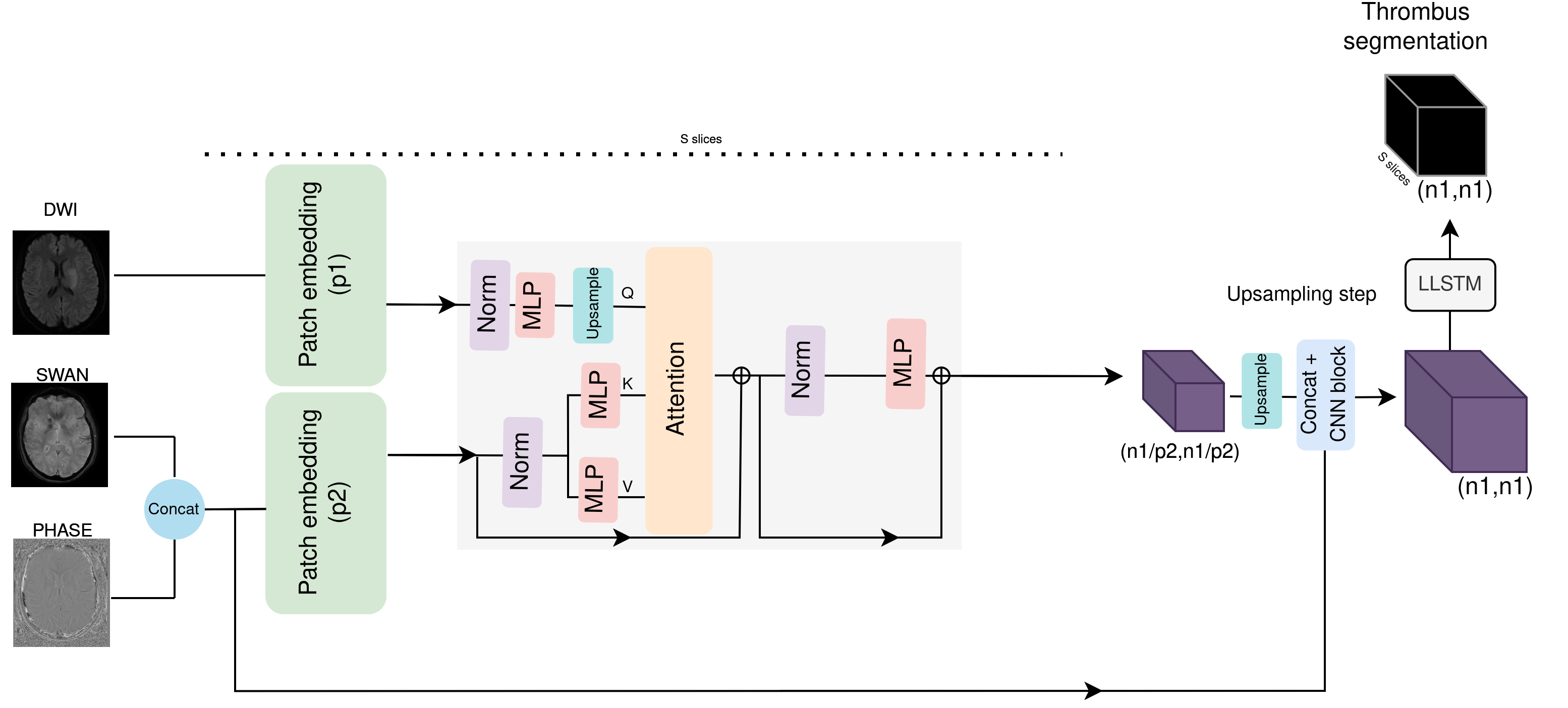}
 \caption{UpAttLLSTM. The thrombus is segmented using a cross-attention module between DWI, SWAN and PHASE, followed by a residual connection with the susceptibility input images and by LLSTM architecture. \label{fig:over}}
 \end{figure*}

In the stroke detection process, the neurologists look for the lesion first. Being a bigger object, it is useful information for detecting the thrombi. As the artery occlusion is the cause of the injured tissue, there are some distance constraints between the thrombi and lesion location: same hemisphere affected, for example.  Trying to mimic that reasoning, we propose the architecture Upsampled Attention Logic LSTM  (UpAttLLSTM), which consists on  AttLLSTM \cite{sofia}, where an upsampling block is added.

The attention block present in the architecture merges the modalities where the lesion (diffusion modalities) and thrombi (susceptibility modalities) are visible using cross-attention. The upsampling block is defined by a residual connection between the modalities where the thrombi are seen and the features obtained after the attention block, helping to get back some details lost during the cross-attention. The result of these two blocks is fed to  \ac{llstm}, which is an adaptation of CLSTM \cite{ConvLSTM} with fewer trainable parameters. In that recurrent network, the longitudinal direction matches time using $s$ slices. Finally, to introduce the lesion and thrombus distance constrains, some post-processing techniques are proposed.   The details of UpAttLLSTM are shown in Fig.~\ref{fig:over}.

\subsubsection{Cross-attention module (ATT)}
\label{sec:cross}
An attention mechanism tries to prioritize the most relevant zones of input data, which helps the learning process to be more efficient and robust. The attention operator multiplies the input by the computed attention matrix, which ranks the input pixels with values between 0 and 1, allowing to highlight the most informative image regions. 

The attention operation is based on a query ($Q$), a key ($K$), and a value ($V$) with dimension $d_{k}$. With the query and the key, the attention weights are computed and then applied over the values. Mathematically, it is expressed as follows:

\begin{equation}
\mbox{Attention(Q,K,V)} = \mbox{softmax(}\frac{QK^{T}}{\sqrt{d_{k}}})V
\label{eq:att}
\end{equation}

In a  cross-attention case, two inputs are used. $Q$ is calculated from input $I_{1}$ and $K$ and $V$ from input $I_{2}$. Doing that, the attention weights are based on both inputs, but they are applied only over $I_{2}$. 

For the thrombus segmentation, as we want to merge the information from the lesion over the thrombus, we use cross-attention considering DWI as $I_{1}$ and the concatenation of SWAN and PHASE as $I_{2}$. The merged information from the lesion and thrombus visible modalities (DWI, SWAN and PHASE) is applied over the thrombus informative modalities (SWAN and PHASE). Doing that, the search for the thrombi leans on the affected brain region. 

As our inputs are images, the attention block has to be adapted to matrices. This adaptation is denoted as Visual Transformers ViT \cite{vit},  which consists of the following steps: patch embedding, calculation of $Q$, $K$ and $V$ through a normalisation and MLP operator, attention between the queries, keys and values and two residual connections.  
%\begin{figure}[!ht]

%\centering

% \vspace*{-0.1in} 
 %\centering
 %\includegraphics[width=1.0\textwidth]{img/crossattention_final.png}
% \caption{Cross attention block between SWAN, PHASE and DWI. \label{fig:crossatt}}
% \end{figure}

\begin{figure*}[!ht]

\centering

 \includegraphics[width=0.8\textwidth]{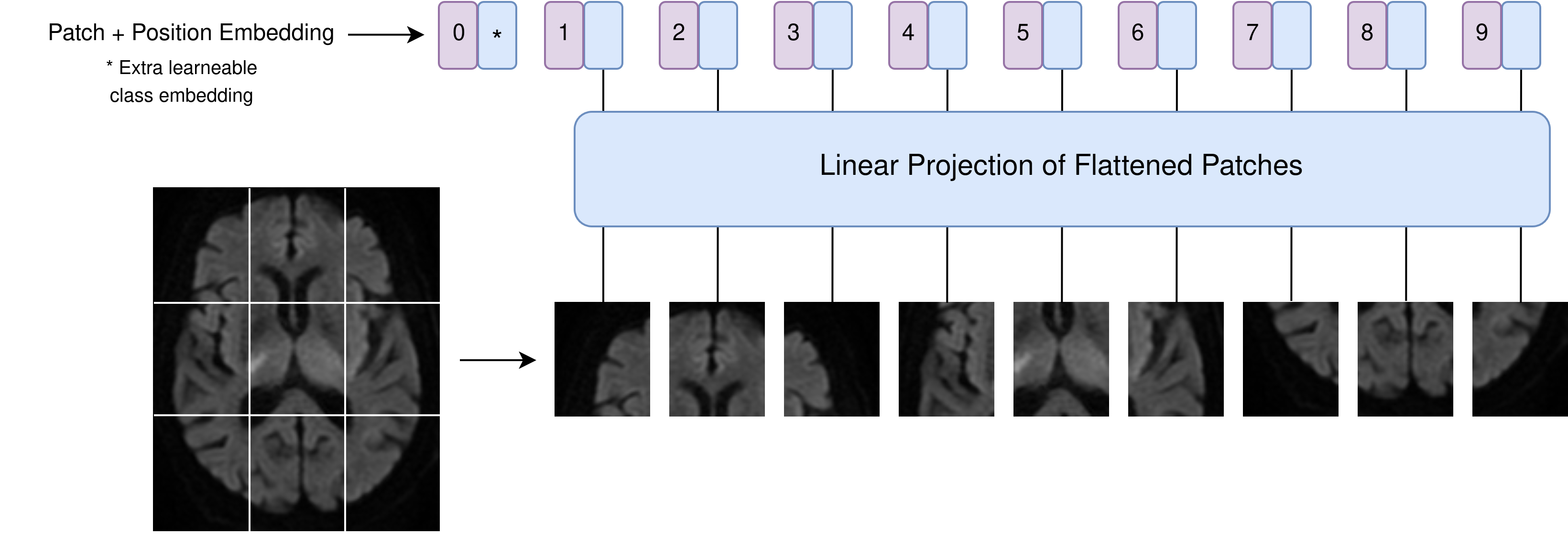}
 \caption{DWI patch embedding in ViT. \label{fig:patchemb}}
 \end{figure*}

The patch embedding operator  is applied over the diffusion and susceptibility modalities. For that, it is calculated per modality using $\mbox{Proj}_{p_{i}}$, which is based on a CNN where the stride and the kernel size are equal, as shown in Fig. \ref{fig:patchemb} This operator produces a linear projection of the flattened patches' sizes, defined by $p_{i}$, and we assume that $d_{k}$ kernels are used. As two projections are done: one over DWI and another over the concatenation of SWAN and PHASE, it produces two outputs: $z^{*}_{1,1}$ and $z^{*}_{1,2}$ with sizes of  $(2\frac{n_{i}}{p_{i}}, d_{k})$. They are calculated as follows:

\begin{equation}
\begin{split}
\mathbf{z}^{*}_{1,1}= &\mbox{Proj}_{p_{1}}(\text{DWI}) \\
\mathbf{z}^{*}_{1,2}= &\mbox{Proj}_{p_{2}}(\text{SWAN}||^4 \text{PHASE}) 
\end{split}
\end{equation}
 In addition, two trainable parameters are added to the projections: a position (Pos) and class (Class) embedding.  The position parameter aims to learn the position information which is lost during the flattening process, and the class one aims to distinguish between different patch classes. These two parameters are concatenated to each output, as in ViT, producing  the following features:

\begin{equation}
\begin{split}
\mathbf{z}_{1,1}= & \text{Class}_{1}||^1 \mathbf{z}^{*}_{1,1}+\text{Pos}_{1}  \\
\mathbf{z}_{1,2}= & \text{Class}_{2} ||^1 \mathbf{z}^{*}_{2,2} +\text{Pos}_{2} 
\end{split}
\end{equation}
The output sizes of the patch embedding operator are $(1+\frac{2n_{1}}{p_{2}},d_{k})$. %A visual schema of DWI embedding process is shown in Fig. \ref{fig:patchemb}. 

These patch embeddings serve for the calculation of $Q$, $K$, and $V$, applying 
a normalisation and an MLP, as follows: 

\begin{equation}
\begin{split}
Q = & \text{MLP}_{1}(\text{Norm}(\mathbf{z}_{1,1})) \\
K= &\text{MLP}_{2}(\text{Norm}(\mathbf{z}_{1,2})) \\
V= &\text{MLP}_{3}(\text{Norm}(\mathbf{z}_{1,2}))
\end{split}
\end{equation}

Using the query, key and value, the attention operation (Eq. \ref{eq:att}) is used, obtaining the output $\mathbf{z}_{2}$. As the attention mask is computed between both types of modalities and  is applied only over the susceptibility modalities, it guides the thrombi search.  In the case where different patch sizes are defined ($p_{1}\neq p_{2}$), there is an upsampled operation needed. 

Assuming that the patch size used for DWI is bigger than the one used for SWAN ($p_{2}<p_{1}$), as the resolution of diffusion modalities are smaller and the observed lesions are bigger, the attention operator becomes:

\begin{equation}
\mathbf{z}_{2} = \text{Attention}(\text{Up}_{\frac{p_{1}}{p_{2}}}(Q),K,V)
\end{equation}

Finally, the output features ($\mathbf{z}_{4}$) are obtained using the following two residual connections:  

\begin{equation}
\begin{split}
\mathbf{z}_{3} = &\text{Norm}(\mathbf{z}_{1,2}[1:]+\lambda_{1}\mathbf{z}_{2}[1:]) \\
\mathbf{z}^{*}_{4} = &\mathbf{z}_{3}+\lambda_{2}\text{MLP}_{5}(\text{Norm}(\mathbf{z}_{3})) \\
\mathbf{z}_{4}=&\text{reshape}(\mathbf{z}^{*}_{4})
\end{split}
\end{equation}

where [1:] represents that the class embeddings are skiped, going back to a size of size ($\frac{2n_{1}}{p_{2}},d_{k}$) and $\lambda_{1}, \lambda_{2}$ are trainable parameters. The final output features  are reshaped to get back an image, obtaining the following output size ($\frac{n_{1}}{p_{2}},\frac{n_{1}}{p_{2}},d_{k}$)

In conclusion, the defined cross-attention block merges the modalities using a specific strategy: to use the lesion information as support and guide for the thrombi location and the features obtained as output are reduced by $p_{2}$ in X  and Y axes. 

\begin{figure*}[!htb]
\vspace*{-0.05in}
 
 \includegraphics[width=1.0\textwidth]{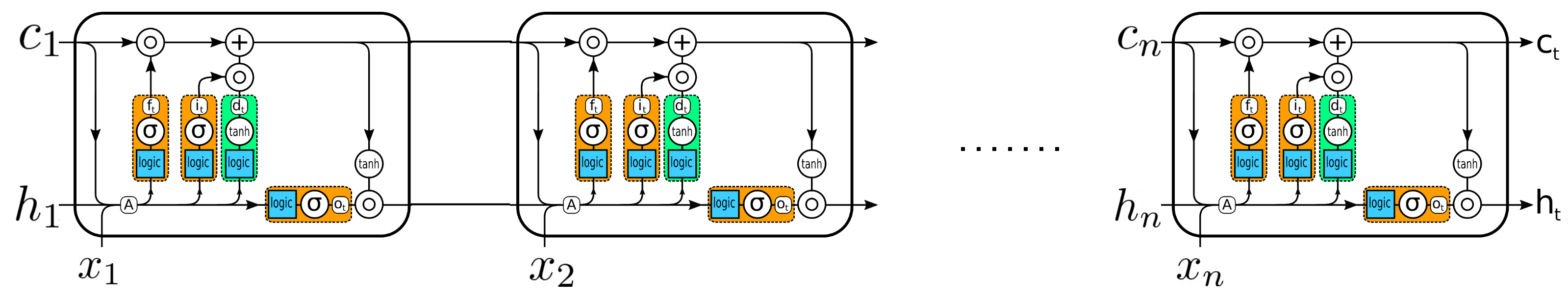}
 \centering
 \caption{LLSTM. The convolution operation is replaced by the Logic one. The architecture produces the output in a recurrent way. \label{fig:LLSTM1}} 
\end{figure*}

\subsubsection{Upsampling (Up)}
\label{sec:up}

After the cross-attention block, the feature resolution is reduced due to the patch embedding. Indeed, the size has been reduced by $p_{2}$ in X and Y. The Upsampling block serves to augment the feature size before getting the segmentation of each input pixel.  We propose to upsample the image using a residual connection with the susceptibility images, to get back some possible tissue details lost during the patch embedding process.

This residual connection concatenates some new features obtained from  SWAN and PHASE to the features obtained in the previous block ($\mathbf{z}_{4}$). First, a convolution is applied over the susceptibility inputs as a non-linear Relu activation function. Their concatenation with the upsampled features are then passed by another convolution and a non-linear elu function with a normalisation layer, as shown in the following equation:

\begin{equation}
\begin{split}
 \mathbf{a}_{1}  = \mbox{ } & \text{Relu}(\mathcal{L}_{1}(\text{SWAN}||^{3}\text{PHASE}))\\
 \mathbf{a}_{2} \mbox{ } = \mbox{ } & \text{Up}_{p_{2}}(\mathbf{z}_{4})\\
  \mathbf{z}_{5} \mbox{ } = \mbox{ } & \text{Elu}(\text{norm}(\mathcal{L}_{2}(\mathbf{a}_{1}||^{3}\mathbf{a}_{2})))
\label{eq:gradmod3}
\end{split}
\end{equation}

The kernel sizes of $\mathcal{L}_{1}$ are of size 3$\times$3, using 12 kernels and of  $\mathcal{L}_{2}$ of size 7$\times$7, using $d_{k}$ kernels. In this case, the convolutional networks are defined with the needed padding size to obtain equal input and output sizes. $\mathbf{z}_{5}$ has the same size as the input size in X-Y dimensions, being $(n_{1}, n_{1}, d_{k})$.

\subsubsection{Logic LSTM (LLSTM)}
\label{sec:llstm}

After the attention and upsampling block, the features obtained represent the susceptibility images merged with the diffusion modalities information. All the previous operations are based on 2D operations, whereas the MRIs inputs are 3D. Applying the cross attention and upsampling method over the total number of  slices ($n_{3}$) produces an output of size  $(n_{1}, n_{1}, n_{3}, d_{k})$ assuming that the image sizes in X and Y are equal ($n_{1}$). We propose to use LLSTM \cite{thesis} over these features for the final segmentation of the thrombus.  

LLSTM, being an adaptation of CLSTM, has fewer parameters than its competitors. As  the attention operator,  due to the several MLPs defined, requires thousands of parameters, a lighter architecture reduces the risk of overfitting. In addition, the recurrency aspect allows to mimic the doctor detection procedure: checking several previous and next slices to detect the object. The reduction of the number of parameters is done by replacing the convolutional operations ($\mathcal{L}(A)$) present in CLSTM by $Logic(A)$, which also increases the receptive field, as shown in Figure \ref{fig:LLSTM1}.

\begin{figure}[!htb]
\vspace*{-0.05in}
 \centering

\includegraphics[width=0.4\textwidth]{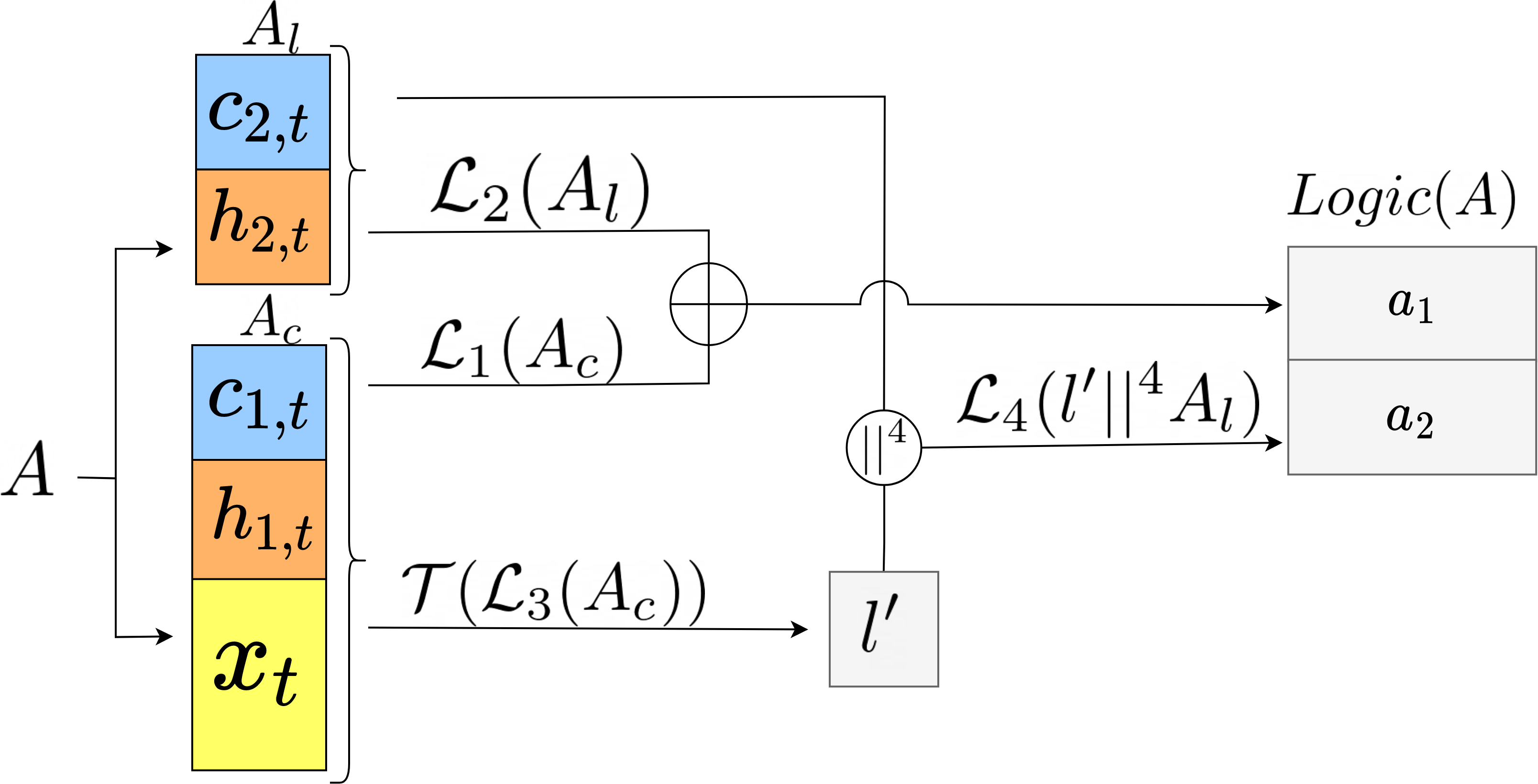}
 \caption{$Logic$ operator. Splitting the input into $A_{l}$ and $A_{c}$, where the convolutions are only applied over $A_{c}$ and a pooling operator $\mathcal{T}$ is used over$A_{l}$, the number of trainable parameters is reduced.  \label{fig:LLSTM2}} 
\end{figure}

 The operation is reduced to the concatenation of a convolution part ($\mathbf{a}_{1}$) and a logic part ($\mathbf{a}_{2}$). The convolution operation is only applied on a part of the input ($A_{c}$), considerably reducing the number of learned parameters. To do so, we first slice $\mathbf{h}_{t}$ and $\mathbf{c}_{t}$ in $\mathbf{h}_{1,t}, \mathbf{h}_{2,t}$ and $\mathbf{c}_{1,t}, \mathbf{c}_{2,t}$. 
\begin{equation}
\begin{split}
 \mathbf{c_{t}} =& \mathbf{c}_{1,t}||^{4}\mathbf{c}_{2,t} \\
 \mathbf{h_{t}} =& \mathbf{h}_{1,t}||^{4}\mathbf{h}_{2,t},
\end{split}
\end{equation}
i.e. the hidden state $\mathbf{h}_t$ is split into a convolution part $\mathbf{h}_{1,t}$ with $n_{c}$ channels and a logic part with $n_{l}$ channels and likewise for the cell state $\mathbf{c}_t$. Using them, we define the splits of $ A_{c}$ and $ A_{l}$ as following:
\begin{equation}
\begin{split}
 A_{c} =& \mathbf{c}_{1,t}||^{4}\mathbf{h}_{1,t}||^{4}\mathbf{x}_{t}  \\
 A_{l} =& \mathbf{c}_{2,t}||^{4}\mathbf{h}_{2,t}.
 \end{split}
\end{equation}
%This repartition serves only organizational purposes so that we know the storage location of convolutional and logical information. The convolutional part is almost identical to the original convolutional \ac{lstm} and serves to store local information, texture for example. The logic part stores information about distant features of larger neighborhoods up to the whole image size. 
Considering that $\mathcal{L}_{i}$ are convolution layers with weights $ W_{i}$ and $b_{i}$ and in particular $\mathcal{L}_{2}$ has $b_{2}= 0$, we obtain the convolution ($\mathbf{a}_{1}$) and the logic ($\mathbf{a}_{l}$) part and $Logic(A)$ as following:
\begin{equation}
\begin{split}
 \mathbf{a_{1}}=&  \mathcal{L}_{1}(A_{c})+\mathcal{L}_{2}(A_{l}) \\\
 \mathbf{a_{2}} =& \mathcal{L}_{4}(\mathcal{T}(\mathcal{L}_{3}(A_{c}))||^{4}A_{l}) \\
Logic(A) =& \mathbf{a}_{1} ||^{4}\mathbf{a}_{2} 
\end{split}
\end{equation} 
where $\mathcal{T}$ is put for the Transfer Layer, which produces the same output size as the input size. $\mathcal{L}_{1},\mathcal{L}_{3}$ and $\mathcal{L}_{2},\mathcal{L}_{4}$ are $w\times w$ and $1\times 1$ convolutions respectively. Finally, denoting $(V_{i_1 i_2 i_3 i_4})_{i_k \in [r_1 \dots r_2]}$ as the slicing $V$ by setting the dimension $i_k$ to a specific range $r_1,\dots,r_2$ or a single value $r_1$, $\mathcal{T}$ is defined as follows:
%p &= \frac{2n_1}{2^i}\\
\begin{equation}\label{transferFunction}
 \left(\mathcal{T}(I)_{i_1 i_2 i_3 i_4} \right)_{i_4 = u} = \text{maxpool}\left((I_{i_1 i_2 i_3 i_4} )_{i_4 = u},p\right)
\end{equation}
where $ u = im+j$ with $i \in \{1,\dots,k\},j \in\{1,\dots,m\}$. $I$ is of dimensions $n_{1},\ n_{2},\ n_{3}=1,\ n_{4}=n_{4_l} = k m$ where the multiplicity $m$ is a positive integer. We choose $p$ as powers of 2 using $p = \frac{2n_1}{2^i}$. For example, having $ n_{4}=6$ and $m=3$, we have $k= 2$, so we apply a pooling layer on the first three features, and another one on the last three ones, each one with different sizes. So, this layer performs a max pooling \cite{Maxpool} operation on each channel separately, with varying window sizes ($p$). A summary of the $Logic$ operator is shown in Fig. \ref{fig:LLSTM2}

The convolution result ($a_{1}$) can be seen as the original \ac{clstm}, being applied to part of $A$ ($ \mathcal{L}_{1}(A_{c})$) and receiving information from the logical part ($ \mathcal{L}_{2}(A_{l})$) as additional bias. %The logical part $c_l, h_l$ is created from the $c_c, h_c$ and the input image in equations \ref{LogicEquation1} and \ref{LogicEquationsEnd}. 
The $\mathcal{T}$ function moves features across the image in a series of different neighbourhood sizes, from local to global. % and this information is stored within the logical part. 
This allows the receptive field of the gates to be as large as the whole input image (which allows doing the concatenation with $A_{l}$) and captures the spatial distance as well. In addition, a double pass is done to not starting with zero memory and to replicate a bidirectional recurrence method but without adding parameters. The first pass is done to save all the memory states, so the prediction is done using them in the second one \cite{thesis}. Finally, before the prediction, a convolutional layer is included of size 1$\times$1 with output channels 2 (number of classes).

Our proposed architecture, UpAttLLSTM, motivated by the neurologist procedure, merges the lesion and thrombus information and recurrently uses the slices, allowing it to produce a robust output by checking previous and future slices based on the coronal MRI space. 
\subsubsection{Post-processing using the lesion information}
\label{sec:post}

Due to the big size of the MRI scans in comparison to the thrombus size, we propose some post-processing techniques to reduce the number of false positives. These techniques are based on two main assumptions: the clot is a dense object and its distance from the lesion is limited.

An artefact, as it does not represent a real object but a noise due to the MRI operations, should be less dense than a real tissue present in the brain. We propose two different approaches: to eliminate all predicted objects that are smaller than $N_{\text{pixels}}$ pixels and to eliminate the predicted objects that are considerably smaller with respect to the biggest one (Big). These two approaches try to prioritise the bigger predictions.

In addition, we propose to use the distance between the lesion and the thrombi predicted masks to improve the model performance, as we know that it exist some biological constraints concerning their distances. To eliminate false positives that are considerably far from the affected zone, we compute the Euclidean distance between the center of mass of both masks and select the closest object to the lesion with at least $N_{\text{dist}}$ pixels of distance from the others. So only the objects $i$ for the patient $j$ satisfying the following property are maintained:

\begin{equation}
  |  \text{dist}(t_{i,j},l_{j})-\min_{i}(\text{dist}(t_{i,j},l_{j}))|<N_{\text{dist}}
\end{equation}
where $t_{i,j}$ is the center of mass of the predicted object $i$ of the patient $j$ and $l_{j}$ is the center of mass of the predicted lesion. 

Finally, to improve the performance, the threshold  ($T$) is reduced in the pixels that are around the predicted object, increasing the size of the predicted thrombus. By around, we mean that we enlarge the detected object by the pixels that touch previous predicted ones with a probability higher than the threshold chosen.

\subsection{Metrics}
The metrics used are the following: the Dice score, which measures the overlap between the prediction and the ground truth, the average count and size of false positives (FP) and false negatives (FN), all of them calculated pixel-wise.  The detection rate (det.) is also calculated by patients, being one if at least one pixel's prediction overlaps the ground truth. 

\section{Results}

We evaluate first our two proposals and compare them with several state of the art methods. Gradual modality dropout is compared with the other model-agnostics methods using nnUnet benchmark with ISLES, CHSF, MATAR and JHU datasets in Subsection  \ref{sec:misssota}. Then, each module proposed for UpAttLLLSTM is evaluated in Subsection \ref{sec:sotath} using CHSF, MATAR and MATAR2 datasets and finally both methods are combined in Subsection \ref{sec:missth}, where FOCH dataset is included in the evaluation. 
\begin{table*}[!htb]
\centering
 \scalebox{0.85}{ \begin{tabular}{cccc|ccccc}
Exp. &\multicolumn{2}{c}{Mod. drop}&ADC &\multicolumn{5}{c}{Test set (T$_{0}$) (Dice)} \\
num.&Type& 1-$p$& & CHSF &MATAR & ISLES & JHU & Average \\
\midrule
Exp1& - &-&\fullcirc& 0.793$\pm$0.030 & 0.787$\pm$0.048&0.739$\pm$0.032 & \textit{0.722$\pm$0.013}& \textit{0.760$\pm$0.047}\\
Exp1&-&-&\emptycirc&0.791$\pm$0.030 & 0.613$\pm$0.067 &0.687$\pm$0.037& 0.653$\pm$0.016& 0.686$\pm$0.059\\
Exp2 &MultiUnet& 0.2&\fullcirc& \textit{0.799$\pm$0.029}&0.764$\pm$0.051& 0.720$\pm$0.033 & 0.714$\pm$0.013 & 0.749$\pm$0.049\\
Exp2 &MultiUnet& 0.2&\emptycirc&\ubold{0.797$\pm$0.030}&0.751$\pm$0.052&0.727$\pm$0.032 &0.695$\pm$0.015&0.743$\pm$0.051 \\ 
Exp3&Gradual & 0.2&\fullcirc&0.776$\pm$0.028 & 0.778$\pm$0.055&0.748$\pm$0.032 & 0.721$\pm$0.013&0.756$\pm$0.049\\
Exp3&Gradual & 0.2&\emptycirc& 0.787$\pm$0.026 & 0.754$\pm$0.058 &0.749$\pm$0.032 & 0.698$\pm$0.015 & 0.745$\pm$0.052 \\
Exp4 &ModDrop+& 0.2&\fullcirc& 0.787$\pm$0.027 & 0.781$\pm$0.051&0.749$\pm$0.032 &0.706$\pm$0.014&0.756$\pm$0.049\\
Exp4 &ModDrop+& 0.2&\emptycirc&0.786$\pm$0.028&0.774$\pm$0.050&0.751$\pm$0.032&0.701$\pm$0.015&0.753$\pm$0.050 \\
Exp5&MultiUnet& 0.5&\fullcirc&0.779$\pm$0.031 & \textit{0.788$\pm$0.049}&\textit{0.759$\pm$0.030} & 0.714$\pm$0.013 & \textit{0.760$\pm$0.047}\\
Exp5&MultiUnet& 0.5&\emptycirc&0.776$\pm$0.031 &\ubold{0.788$\pm$0.048} &\ubold{0.760$\pm$0.031} & 0.709$\pm$0.014 & \ubold{0.758}$\pm$0.048\\ 
Exp6 &Gradual & 0.5&\fullcirc& 0.781$\pm$0.029 & 0.759$\pm$0.060 &0.730$\pm$0.032 &0.713$\pm$0.013 & 0.746$\pm$0.050\\
Exp6&Gradual & 0.5&\emptycirc&0.781$\pm$0.029&0.739$\pm$0.060 & 0.718$\pm$0.035 & 0.692$\pm$0.015& 0.733$\pm$0.054\\
Exp7 &ModDrop+& 0.5&\fullcirc& 0.783$\pm$0.030&0.786$\pm$0.043 &0.744$\pm$0.031& 0.699$\pm$0.014&0.753$\pm$0.047 \\
Exp7 &ModDrop+& 0.5&\emptycirc& 0.779$\pm$0.031 & 0.778$\pm$0.047 &0.736$\pm$0.032 &0.694$\pm$0.015 &0.747$\pm$0.050\\
Exp8 & MultiUnet& 0.8&\fullcirc& 0.787$\pm$0.029 &0.768$\pm$0.058&0.751$\pm$0.031 &0.706$\pm$0.014 & 0.753$\pm$0.051\\
Exp8 &Full & 0.8&\emptycirc& 0.787$\pm$0.029&0.764$\pm$0.059& 0.753$\pm$0.032 & 0.706$\pm$0.014 & 0.753$\pm$0.051\\
Exp9 &Gradual& 0.8&\fullcirc& 0.789$\pm$0.029 &0.771$\pm$0.049 &0.731$\pm$0.032 & 0.712$\pm$0.013 & 0.751$\pm$0.048 \\
Exp9 &Gradual & 0.8&\emptycirc& 0.788$\pm$0.029&0.752$\pm$0.049&0.729$\pm$0.032&0.691$\pm$0.015 & 0.740$\pm$0.050\\
Exp10 &ModDrop+& 0.8&\fullcirc& 0.775$\pm$0.031&0.768$\pm$0.052&0.753$\pm$0.031 &0.715$\pm$0.014 &0.753$\pm$0.049\\
Exp10 &ModDrop+& 0.8&\emptycirc& 0.773$\pm$0.032& 0.767$\pm$0.052 &0.754$\pm$0.031& \ubold{0.712$\pm$0.014}&0.752$\pm$0.049\\
\bottomrule
\end{tabular}}
 \caption{Results for \ac{adc}-missing evaluations using modality dropout with $p$ and either gradual (grad. \cmark) or not: 
 The training uses all datasets with \ac{adc} and \ac{dwi} inputs. The best Dice score per dataset and the best overall average are highlighted. Evaluation symbols: $\fullcirc$ represents all modalities present, while $\emptycirc$ indicates \ac{adc} is missing. \label{tab:miss1}}
%\label{tab:tab22}
\end{table*}

\subsection{Gradual modality dropout: SOTA comparison}
\label{sec:misssota}

The model's robustness to missing modalities is assessed in two scenarios: (1) in-domain, where all datasets are included in the training, and (2) out-domain, where evaluation is conducted on an unseen center.  Only techniques compatible with nnUnet benchmark training are compared with our proposed method. Specifically, we evaluate MultiUnet~\cite{dropmultidata}, where the gradual function $g(t)$ remains fixed at zero through all iterations, and ModDrop+~\cite{moddropplus}, which includes a dynamic filter head before the first layer of nnUnet that changes depending on the input position of the absent modality if there is one. Approaches based on GANs or reconstruction methods are excluded, as they would require architectural modifications. 

\subsubsection{In domain evaluation}

Using \ac{dwi} and \ac{adc} as inputs and the four datasets for training in an in-domain framework, the dropout is applied exclusively to ADC by setting different $p$ values in Equation \ref{eq:gradmod}, ensuring that DWI remains present throughout training. The model is evaluated with all modalities present and under simulated missing-data conditions. Results are summarised in Table \ref{tab:miss1}.

Performance significantly drops when \ac{adc} is missing, particularly for MATAR, with the average Dice score decreasing from 0.787 to 0.613 (Exp1, first vs. second row). In contrast, when applying modality dropout, either gradual or not, and with the dynamic filter of ModDrop+ performance is stable (Exp2-10): the Dice scores are maintained between 0.733 and 0.760 (Exp2-10, second row) while preserving near-optimal performance when all modalities are available (0.746–0.760, Exp2-10, first row). The best average Dice (0.760) is achieved with $(1-p)=0.5 $ for MultiUnet (Exp5, first row) but in the case of gradual dropout with $(1-p)=0.2$  (Exp3, first row) or ModDrop+ with $(1-p)=0.2$ the Dice arrives to 0.756, being close to the best score. When \ac{adc} is missing, the best performance is obtained with MultiUnet (Dice = 0.758 Exp5 second row) and with either ModDrop+ or gradual modality dropout, a maximum Dice of 0.753 is obtained. All these results arrives to a comparable performance of the best scenario where all modalities were used. In conclusion, applying any of the methods does not degrade the performance when all modalities are available and minimises loss when one modality is missing, all achieving comparable results.

\subsubsection{Out domain evaluation}

\begin{table*}[!htb]
\centering

\scalebox{0.85}{ 
\begin{tabular}{ccccc|cccccc}
Exp. &\multicolumn{2}{c}{Mod. drop}&\multicolumn{1}{c}{B0} & Unseen &\multicolumn{4}{c}{Test set (T$_{0}$) (Dice)}\\
num &Type & 1-$p$ & & ISLES& CHSF &MATAR &ISLES (unseen) & JHU &Average \\
\midrule
Exp11&-&- & $\fullcirc$& \cmark&0.790$\pm$ 0.029 & 0.768$\pm$0.061&- & 0.731$\pm$0.012&0.763$\pm$0.059\\

Exp11&-&- &$\emptycirc$& \cmark& 0.787$\pm$0.027 & 0.767$\pm$0.065 & $0.622^{ \dagger}\pm$ 0.040 & 0.723$\pm$0.014& 0.725$\pm$0.055\\
%%% DONEEE

Exp12&MultiUnet&0.2 &$\fullcirc$ & \cmark&0.787$\pm$0.027 &0.762$\pm$0.056 & -&0.750$\pm$0.012&0.767$\pm$0.061\\

Exp12&MultiUnet&0.2 &$\emptycirc$& \cmark&0.788$\pm$0.028 &0.765$\pm$0.056 & 0.672$\pm$0.039 & 0.748$\pm$0.013 &0.743$\pm$0.051\\

Exp13&Gradual&0.2 & $\fullcirc$& \cmark&0.789$\pm$0.028& 0.757$\pm$0.061& - & 0.744$\pm$0.013&0.763$\pm$0.061\\

Exp13&Gradual&0.2&$\emptycirc$& \cmark& 0.789$\pm$0.028 & 0.758$\pm$0.061 & 0.650$\pm$0.040 &0.743$\pm$0.013 & 0.735$\pm$0.052\\

Exp14&ModDrop+&0.2 &$\fullcirc$& \cmark &0.787$\pm$0.028&0.733$\pm$0.066& - & 0.740$\pm$0.014&0.753$\pm$0.066\\
Exp14&ModDrop+&0.2 &$\emptycirc$& \cmark&0.787$\pm$0.029 &0.732$\pm$0.065&0.617$\pm$0.041&0.740$\pm$0.014&0.719$\pm$0.056\\

Exp15&MultiUnet&0.5 &$\fullcirc$& \cmark& 0.789$\pm$0.028 & $\underline{0.775\pm0.056}$ & - &0.748$\pm$0.013 &$\underline{0.771\pm0.058}$\\
Exp15&MultiUnet&0.5 &$\emptycirc$& \cmark& 0.789$\pm$0.028&$\ubold{0.777\pm0.055}$ & 0.637$\pm$0.040&0.747$\pm$0.013 & 0.738$\pm$0.051\\ 

Exp16&Gradual&0.5 & $\fullcirc$& \cmark&0.786$\pm$0.029& 0.766$\pm$0.055& - & 0.747$\pm$0.013 &0.766$\pm$0.058\\
Exp16&Gradual&0.5&$\emptycirc$& \cmark& 0.786$\pm$0.029 & 0.763$\pm$0.056 & 0.655$\pm$0.039 &0.748$\pm$0.012 & 0.738$\pm$0.050\\
%##AQUI AHORA
%########################

Exp17&ModDrop+&0.5 &$\fullcirc$& \cmark &0.779$\pm$0.029&0.758$\pm$0.070& - & 0.744$\pm$0.013&0.760$\pm$0.066\\
Exp17&ModDrop+&0.5 &$\emptycirc$& \cmark&0.778$\pm$0.029 &0.761$\pm$0.070& 0.630$\pm$0.040&0.743$\pm$0.013&0.728$\pm$0.056\\
%AQUI AHORA
%#############################
Exp18&MultiUnet&0.8 & $\fullcirc$& \cmark&0.790$\pm$0.026 & 0.743$\pm$0.068&- &0.751$\pm$0.013 &0.761$\pm$0.065\\
Exp18&MultiUnet&0.8 & $\emptycirc$& \cmark&0.790$\pm$0.027 & 0.741$\pm$0.068 & 0.682$\pm$0.032&0.751$\pm$0.013 & 0.741$\pm$0.050\\ 

Exp19&Gradual&0.8 &$\fullcirc$& \cmark & $\underline{0.810\pm0.025}$ & 0.750$\pm$0.055& - & 0.743$\pm$0.013 & 0.768$\pm$0.057\\
Exp19&Gradual&0.8 &$\emptycirc$& \cmark& $\ubold{0.810\pm0.026} $& 0.752$\pm$0.056 & $\ubold{0.701}\pm0.038$ &0.743$\pm$0.013 &$\ubold{0.749}\pm0.048$\\

Exp20&ModDrop+&0.8 &$\fullcirc$& \cmark &0.788$\pm$0.026&0.763$\pm$0.060& - & $\underline{0.753\pm0.012}$&0.768$\pm$0.058\\
Exp20&ModDrop+&0.8 &$\emptycirc$& \cmark&0.788$\pm$0.027 &0.761$\pm$0.060& 0.635$\pm$0.038&$\ubold{0.753\pm0.012}$&0.734$\pm$0.051\\
Exp21&-&-& $\fullcirc$& &0.773$\pm$0.031&0.753$\pm$0.056&- &0.756$\pm$0.012 &0.761$\pm$0.057 \\

Exp21&-&-& $\emptycirc$&& 0.775$\pm$0.025&0.680$\pm$0.058&$0.782^{*}\pm$0.029 &0.756$\pm$0.012 & 0.747$\pm$0.041 \\
\bottomrule
\end{tabular}}
 \caption[SOTA missing modality results under an out-domain scenario]{ 
 The evaluation considers missing B0 on the unseen ISLES dataset, using modality dropout with probability $p$ in gradual and non-gradual settings. All training configurations use ADC, DWI and B0 as inputs. The upper bound (*) includes ISLES in training, while the lower bound ($\dagger$) excludes it, both without any method. The best Dice scores for unseen datasets and average results under missing-modality conditions are highlighted and the best without absent data are underlined. Evaluation symbols: $\fullcirc$ for all modalities present, and $\emptycirc$ when B0 is absent. \label{tab:res3}}%
\end{table*}

\begin{figure}[!htb]
 \centering
\includegraphics[width=0.45\textwidth]{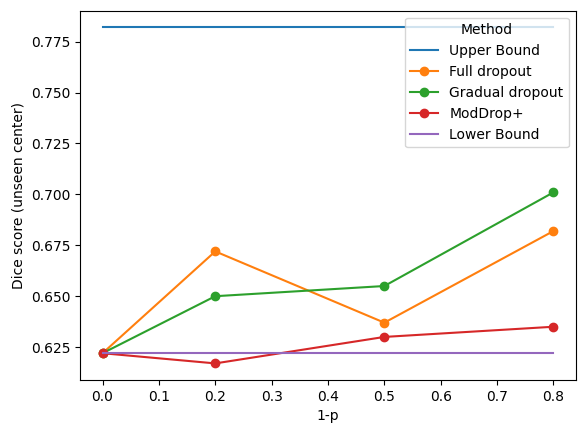}
 \caption{Dice score results on unseen center (ISLES) varying $p$ using nnUnet with ADC, DWI, and a black image instead of B0 as inputs. The upper bound is when ISLES is used during training, and the lower bound is when no method is used. \label{fig:dropoutcurve}}
\end{figure}

In real-world scenarios, missing modalities often result from differences in imaging equipment specifications or the absence of specific imaging machines during training. ISLES (which lacks \ac{b0}) is excluded from training to simulate this. Using \ac{adc}, \ac{dwi}, and \ac{b0} as inputs, we apply modality dropout on \ac{b0} with varying probabilities and evaluate performance with all modalities available and under missing-data conditions (B0 for ISLES). Results are shown in Table \ref{tab:res3}, and Fig. \ref{fig:dropoutcurve} shows how the performance on ISLES changes depending on $p$. The lower bound corresponds to training without ISLES and no dropout (Exp11, Dice = 0.622 in ISLES), while the upper bound includes ISLES during training with a black image instead of B0 input (Exp21, Dice = 0.782).

All evaluated methods --gradual modality dropout, MultiUnet and ModDrop+-- demonstrate improved generalisation on the unseen dataset compared to the lower bound, except for MultiUnet with $(1-p)$=0.2 (Exp14).  These approaches achieve Dice scores ranging from 0.637 and 0.701 (Exp12-20, second-row ISLES, except for Exp14) while preserving performance across other datasets. Additionally, they stabilise results when \ac{b0} is absent, maintaining consistent Dice scores (Exp12-20, first vs. second-row comparison). The best performance on the unseen dataset is obtained with gradual modality dropout at $(1-p) = 0.8$, reaching a Dice score of 0.701 (Exp19). This approach effectively leverages data augmentation, allowing nnUnet to generalise better across different medical centers. In contrast, MultiUnet results in lower performance (Dice=0.682 in Exp18), and introducing a dynamic filter head via ModDrop+ further reduces accuracy (Dice=0.635 in Exp20). 

Our proposed method also improves segmentation on other datasets, yielding the highest Dice scores for CHSF, regardless of \ac{b0} availability (Exp19, 0.810 Dice score in both first and second rows). On average, performance remains stable when all modalities are present (first rows of Exp11 vs Exp12-20) and gets better results in comparison to the other methods when B0 is absent (second rows of Exp11 vs Exp12-20). Specifically, gradual modality dropout enables Dice scores close to the upper bound (approximately 0.75 in Exp19, second row).

%In Fig. \ref{fig:lesionisles}, the predictions of several models and the groundtruth are represented over the DWI image for ISLES dataset. All models have DWI, ADC, and missing B0 as input (black image). For the first column, the lower bound model is used; when ISLES is not seen during training, the second one is the best model obtained by adding the proposed method ($(1-p)$=0.8), and the third column is when the model uses ISLES during training. Finally, in the last column, the groundtruth of the model is shown. In the first line, when the dataset is unseen and no method to manage the missing modality is used, some parts of the lesion are not detected. For the second case, false positives are obtained if the method is not applied. Finally, in the last case, we can see how seeing or not seeing ISLES, there are parts of the lesion that are still not detected. 

In conclusion, our method most effectively strengthens nnUNet’s ability to process unseen datasets with missing modalities while ensuring no degradation in performance on previously encountered datasets.  

%\begin{figure}[!htbp]

% \begin{tabular}{@{}c@{ }c@{ }c@{ }c@{}c@{ }}
%Unseen ISLES & Unseen ISLES & Seen ISLES & Groundtruth\\
%& + grad. mod. drop.& \\
%\includegraphics[width=0.22\textwidth]{img/nomoddrop.png}&
%\hspace*{0.05in}
%\includegraphics[width=0.22\textwidth]{img/moddrop.png}&
%\hspace*{0.05in}
%\includegraphics[width=0.22\textwidth]{img/bestdrop.png}&
%\hspace*{0.02in}
% \includegraphics[width=0.22\textwidth]{img/gt_islestest.png}\\

%\includegraphics[width=0.22\textwidth]{img/nomoddrop2.png}&
%\hspace*{0.05in}
%\includegraphics[width=0.22\textwidth]{img/moddrop2.png}&
%\hspace*{0.05in}
%\includegraphics[width=0.22\textwidth]{img/bestdrop2.png}&
%\hspace*{0.02in}
%\includegraphics[width=0.22\textwidth]{img/gt_islestest2.png} \\

%\includegraphics[width=0.22\textwidth]{img/nomoddrop3.png}&
%\hspace*{0.05in}
%\includegraphics[width=0.22\textwidth]{img/moddrop3.png}&
%\hspace*{0.05in}
%\includegraphics[width=0.22\textwidth]{img/bestdrop3.png}&
%\hspace*{0.02in}
%\includegraphics[width=0.22\textwidth]{img/gt_islestest3.png}
%\end{tabular}

%\caption{Lesion segmentation from ISLES dataset using ADC, DWI and missing B0 (black image) %\label{fig:lesionisles}}
%\end{figure}
%\newpage

\subsection{Monocentric thrombi segmentation}

For the thrombus segmentation, we have proposed an specific architecture denoted as UpAttLLSTM. To test its performance, only the datasets where the thrombi are annotated can be used: being CHSF, MATAR, MATAR2 and FOCH. As the proposed architecture uses \ac{dwi}, \ac{swan} and PHASE, FOCH is excluded for now, as PHASE is absent and the missing modality problem should be managed. 

%First, we compare several SOTA methods (nnUnet and CLSTM) with our proposed architecture using the Dice score, detection rate, false positives and false negatives in Subsection \ref{sec:sotath}, measuring the impact and advantages of each module proposed using CHSF and MATAR.  Finally, in Subsection \ref{sec:up}, we test the incorporation of Upsampling block and the inclusion of MATAR2 dataset. 

The dataset is normalised using the histogram references per modality using Nyul normalization \cite{nyul} using CHSF and MATAR patients for the standard histogram computation.  The skull-stripped \ac{mri}s are used and DWI is coregistered to SWAN using ANTS software, producing images of size 512$\times$512 in the ($x$,$y$)-plane and $z$ varies between patients. 

As input for the model, we take  256$\times$256$\times s$-size crops, where $s$ is the chosen number of slices. To reduce $x$ and $y$, the center of mass is calculated to add around 128 pixels in all dimensions (arriving at a size of 256$\times$256 around that center). As DWI and SWAN have different resolutions, and indeed the lesion is normally a bigger region than the thrombi, the crops in $z$ are done in the original resolution, taking the corresponding slices. Having $s$ slices from SWAN and  $s$ slices from DWI means that we are seeing a bigger brain region in the diffusion modality, which allows the model to have an increased attention impact.

We use flipping (in three dimensions) and Gaussian noise (each one is chosen in every iteration with 40$\%$ of probability) as data augmentation. 
 As the thrombus is a small object, we use per iteration a crop including the thrombi and another without to manage the unbalanced dataset. We have a sequence of slices containing the full thrombus (successively as the thrombus is a dense object) ans the previous and next slices (all of them without thrombi). So per iteration, one crop of $s$ slices is chosen from the first group of crops and another from the rest.

The training configuration is described in Table \ref{tab:hyp}, where the batch size, the number of crops per image, the patch sizes, the attention embedding ($d_{k}$), the LLSTM hyperparameters ($n_{c}, n_{l}, m, n_{4}$), the proportion of training, validation and test set and the threshold used in the post-processing method ($T$) as well as the minimum number of pixels for the predicted object ($N_{pixels}$) and the minimun distance with the other thrombi concerning the lesion predicted mask ($N_{dist}$). 
\begin{table}[H]

\centering

 \scalebox{0.8}{\begin{tabular}{ccccccccccccc}
 \toprule
 Batch size & Batch's crops & $p_{1}$& $p_{2}$ & $d_{k}$ & $n_{c}$ & $n_{l}$ & $m$& $n_{4}$ & $T$ &  $N_{pixels}$&$N_{dist}$\\
 \midrule
 2 	& 4 & 32& 4 &32 &4 & 8 & 3 &32  & 0.3 & 20 & 20\\
 \bottomrule
 \end{tabular}}
  {\caption{\label{tab:hyp}Training process. Adam optimizer is used with learning rate (lr) 0.01}}
\end{table}

\label{sec:sotath}

We compare each of our proposed modules with some SOTA methods. To do that, just CHSF and MATAR datasets are used during training and the impact of the inclusion of the attention block between \ac{dwi} and \ac{swan} and PHASE and the replacement of the convolution operation by the $Logic$ operator in CLSTM (becoming LLSTM)  are measured. All these proposals are only compared with nnUnet and CLSTM models. The other proposed methods, as they use the hemisphere position \cite{hemisphere} or the reference path through the brainstem as annotations \cite{polarunet} are not included. Also we include the lesion and threshold (Thr) postprocessing techniques, the upsampling block and to choose the biggest object.

%We compute the Dice score, which is calculated pixel-wise and measures the overlap between the prediction and the ground truth, the average count and size of false positives (FP) and false negatives (FN) (in pixel count), and the detection rate (det.) calculated by patients, being one if at least one pixel's prediction overlaps the ground truth. 

First, as the number of slices is a critical hyperparameter on the recurrent methods presented (in nnUnet, the software chooses the crops by itself), a specific study on that parameter is done. The results using several slices are in Table \ref{tab:th1} for proximal (P) and distal (D) occlusions. 

\begin{table}[H]
 \centering

\scalebox{0.8}{\begin{tabular}{cc@{\hspace{-0.1ex}}c@{\hspace{-0.1ex}}cc@{\hspace{-0.1ex}}c@{\hspace{-0.1ex}}cc@{\hspace{-0.1ex}}c@{\hspace{-0.1ex}}cc@{\hspace{-0.1ex}}c@{\hspace{-0.1ex}}cc@{\hspace{-0.1ex}}c@{\hspace{-0.1ex}}cc@{\hspace{-0.1ex}}c@{\hspace{-0.1ex}}c}
\toprule
Slices &\multicolumn{6}{c}{False Positives}& \multicolumn{6}{c}{False Negatives} &\multicolumn{3}{c}{Dice}&\multicolumn{3}{c}{Det. } \\ \hspace{0.1in}
&\multicolumn{3}{c}{Count} & \multicolumn{3}{c}{Size} & \multicolumn{3}{c}{Count} & \multicolumn{3}{c}{Size} & & \\
& P&$\mid$& D& P & $\mid$ & D & P &$\mid$& D &P &$\mid$ & D & P &$\mid$& D & P &$\mid$& D 
\\
\midrule

10 &1.3 &$\mid$ &0.9&74.1&$\mid$ &102.7& 0.3 &$\mid$& 0.3&292.7 &$\mid$& 79.4& 0.48& $\mid$ &0.51 &0.9 &$\mid$ &0.91 \\

 \textbf{12}&\textbf{1.0}& $\mid$& \textbf{0.6}&\textbf{49.8}&$\mid$&\textbf{63.3}&\textbf{ 0.2} &$\mid$& \textbf{0.3}&\textbf{0.2 }&$\mid$ &\textbf{79.4}& \textbf{0.55 }&$\mid$ &\textbf{0.54} &\textbf{1.0}& $\mid$& \textbf{0.91 }\\
14&3.3& $\mid$& 2.3&79.4&$\mid$ &82.8& 0.2 &$\mid$ &0.4&0.2&$\mid$& 108.5& 0.59& $\mid$ &0.42 &1.0& $\mid$& 0.85 \\

\bottomrule
\end{tabular}}
\caption{Results using different numbers of slices for segmenting the thrombi using AttLLSTM with CHSF and MATAR dataset. \label{tab:th1}}
\end{table}

\begin{table*}[!ht]
 \centering
 \scalebox{0.85}{\begin{tabular}{cccccc@{\hspace{-0.1ex}}c@{\hspace{-0.1ex}}cc@{\hspace{-0.1ex}}c@{\hspace{-0.1ex}}cc@{\hspace{-0.1ex}}c@{\hspace{-0.1ex}}cc@{\hspace{-0.1ex}}c@{\hspace{-0.1ex}}cc@{\hspace{-0.1ex}}c@{\hspace{-0.1ex}}cc@{\hspace{-0.1ex}}c@{\hspace{-0.1ex}}cc}
\toprule
Model & Datasets & Lesion & Thr &Big&\multicolumn{6}{c}{False Positives}& \multicolumn{6}{c}{False Negatives} &\multicolumn{3}{c}{Dice}&\multicolumn{3}{c}{Det. } & \# Param \\ \hspace{0.1in}
&&&&& \multicolumn{3}{c}{Count} & \multicolumn{3}{c}{Size} & \multicolumn{3}{c}{Count} & \multicolumn{3}{c}{Size}& & &&&&& (millions) \\
&&&&& P&$\mid$& D& P & $\mid$ & D & P &$\mid$& D &P &$\mid$ & D & P &$\mid$& D & P &$\mid$& D 
\\
\midrule

nnUnet &CHSF+MATAR && & &0.6& $\mid$& 1.1&37.1&$\mid$&222.4& 0.4 &$\mid$& 0.5&373.6&$\mid$&125.3& 0.46& $\mid$ &0.45&0.7 &$\mid$& 0.7& $\simeq$30\\
 
CLSTM & CHSF+MATAR && & &3.8& $\mid$& 2.6&178.1&$\mid$&153.5& 0.2& $\mid$ &0.4&0.2&$\mid$&28.1& 0.39 &$\mid$ &0.33 &1.0 &$\mid$ &0.81 & $\simeq$ 2\\ 

LLSTM & CHSF+MATAR & && &0.8& $\mid$& 1.7&111.9&$\mid$&34.9& 0.2& $\mid$ &0.6&0.2&$\mid$&136.4& 0.48 &$\mid$ &0.36 &1.0 &$\mid$ &0.64& $\simeq$ 1 \\

AttCLSTM &CHSF+MATAR && & &1.2& $\mid$&1.6 &43.81&$\mid$&99.3& 0.3& $\mid$ &0.4&141.1&$\mid$& 100.7& 0.41 &$\mid$ &0.38 &0.9&$\mid$ &0.72 &$\simeq$ 3\\

AttLLSTM & CHSF& &&&3.8 &$\mid$ &3.0&111.9&$\mid$&101.5& 0.2 &$\mid$& 0.6&0.2 &$\mid$&136.4& 0.50 &$\mid$ &0.27 &1.0 &$\mid$& 0.64 & $\simeq$2\\ 
AttLLSTM &CHSF+MATAR & &&&1.0& $\mid$& 0.6&49.8&$\mid$&63.3& 0.2&$\mid$& 0.3&0.2&$\mid$ &79.4& 0.55 &$\mid$ &0.54 &1.0& $\mid$&0.91& $\simeq$ 2 \\
nnUnet &CHSF+MATAR &\checkmark && &0.0& $\mid$& 0.7&0.0&$\mid$&161.3& 0.4 &$\mid$& 0.5&373.6&$\mid$&130.9& 0.48& $\mid$ &0.51&0.7 &$\mid$& 0.7& $\simeq$ 30\\ 
AttCLSTM &CHSF+MATAR &\checkmark && &0.5& $\mid$& 0.9&19.4&$\mid$&109.8& 0.3& $\mid$ &0.6&141.1&$\mid$&134.4 & 0.43 &$\mid$ &0.36&0.9&$\mid$ &0.6 & $\simeq$ 3\\
AttLLSTM &CHSF+MATAR &\checkmark && &0.0 &$\mid$ &0.1&0.0&$\mid$&8.5& 0.2 &$\mid$& 0.3&0.2&$\mid$&79.4& 0.52& $\mid$& 0.58 &1.0& $\mid$ & 0.91 & $\simeq$ 2\\ 
nnUnet &CHSF+MATAR &\checkmark & \checkmark &&0.0& $\mid$& 0.8&0.0&$\mid$&177.9& 0.5 &$\mid$& 0.5&373.6&$\mid$&130.9& 0.49& $\mid$ &0.51&0.7 &$\mid$& 0.7& $\simeq$ 30\\ 
AttCLSTM &CHSF+MATAR &\checkmark & \checkmark& &0.5& $\mid$& 0.9&59.9&$\mid$&245.8& 0.3& $\mid$ &0.6&141.1&$\mid$&134.4 & 0.54&$\mid$ &0.32&0.9&$\mid$ &0.6 & $\simeq$ 3\\
AttLLSTM &CHSF+MATAR &\checkmark &\checkmark &&0.0& $\mid$ &0.1&0.0&$\mid$&8.5& 0.2& $\mid$ &0.3&0.2&$\mid$&79.4& 0.63& $\mid$ &0.59&1.0& $\mid$ &0.91& $\simeq$ 2 \\

AttLLSTM & CHSF+MATAR +MATAR2 & && & 1.0& $\mid$& 1.0&31.7&$\mid$&27.4& 0.3&$\mid$& 0.2&78.1&$\mid$ &62.4& 0.53 &$\mid$ & 0.61 &0.90& $\mid$& 0.93& $\simeq$ 2\\

 \underline{UpAttLLSTM } & CHSF+MATAR+MATAR2 & & && \underline{2.1} & $\mid$&  \underline{2.2}& \underline{34.2}&$\mid$& \underline{56.9}& \underline{0.1} &$\mid$&  \underline{0.2}& \underline{0.1}&$\mid$ & \underline{62.4}& \underline{0.62}&$\mid$ &  \underline{0.58} & \underline{1.00}& $\mid$& \underline{0.93}& $\simeq$ 2\\
\textbf{UpAttLLSTM}  &CHSF+MATAR +MATAR2  && &\checkmark &\textbf{0.3}&  $\mid$& \textbf{0.1}&\textbf{7.8}&$\mid$&\textbf{12.5}&\textbf{0.1} &$\mid$&\textbf{ 0.3}&\textbf{0.1}&$\mid$ &\textbf{65.4}& \textbf{0.65}&$\mid$ & \textbf{0.65} &\textbf{1.00}& $\mid$&\textbf{0.86}& $\simeq $ 2\\
\bottomrule
\end{tabular}}
\caption{Results using different architectures and configurations for segmenting the thrombi using 12 slices for the recurrent methods. The best model is highlighted in bold and the best model without post-processing techniques is underlined. \label{tab:th}}
\end{table*}
Using $s$ equal to 12 for all the recurrent methods, as it produces the best result using AttLLSTM, we compare the results with some of the SOTA methods for the thrombi segmentation in Table \ref{tab:th}. nnUnet is trained using the general procedure, and the following contributions are tested: first, just \ac{llstm} and CLSTM (adding at the beginning a CNN with 5$\times$5 filters instead of the attention), the attention module, and the two post-processing techniques (lesion distance and threshold (Thr)).

%\begin{table*}[!ht]
 %\centering

  %\scalebox{0.6}{\begin{tabular}{cccccc@{\hspace{-0.1ex}}c@{\hspace{-0.1ex}}cc@{\hspace{-0.1ex}}c@{\hspace{-0.1ex}}cc@{\hspace{-0.1ex}}c@{\hspace{-0.1ex}}cc@{\hspace{-0.1ex}}c@{\hspace{-0.1ex}}cc@{\hspace{-0.1ex}}c@{\hspace{-0.1ex}}cc@{\hspace{-0.1ex}}c@{\hspace{-0.1ex}}c}
%\toprule
%Model & Datasets &\multicolumn{3}{c}{Postprocessing}&\multicolumn{6}{c}{False Positives}& \multicolumn{6}{c}{False Negatives} &\multicolumn{3}{c}{Dice}&\multicolumn{3}{c}{Det. } \\ \hspace{0.1in}
%&& Lesion & Thr &Big& \multicolumn{3}{c}{Count} & \multicolumn{3}{c}{Size} & \multicolumn{3}{c}{Count} & \multicolumn{3}{c}{Size} & & \\
%&&&&& P&$\mid$& D& P & $\mid$ & D & P &$\mid$& D &P &$\mid$ & D & P &$\mid$& D & P &$\mid$& D 
%\\
%\midrule
%AttLLSTM &CHSF+MATAR && &&1.0& $\mid$&0.6&49.8&$\mid$&63.3& 0.2&$\mid$& 0.3&0.2&$\mid$ &79.4& 0.55 &$\mid$ &0.54 &1.00& $\mid$& 0.91 \\

%AttLLSTM &CHSF+MATAR &\checkmark &\checkmark&&0.0& $\mid$& 0.1&0.0&$\mid$&8.5& 0.2&$\mid$& 0.3&0.2&$\mid$ &79.4& 0.63 &$\mid$ &0.59 &1.00& $\mid$& 0.93 \\

\begin{figure*}
\centering
\begin{tabular}{c@{}c@{ }c@{ }c@{ }c@{}c@{ }}
 &Groundtruth & CLSTM & LLSTM & \underline{\textbf{AttLLSTM}}\\
 
 & \includegraphics[width=0.16\textwidth]{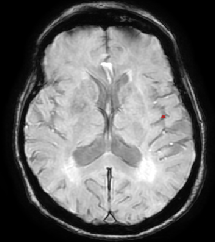} & \includegraphics[width=0.16\textwidth]{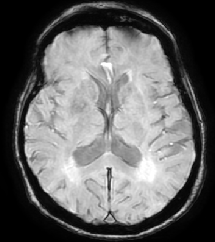} & \includegraphics[width=0.16\textwidth]{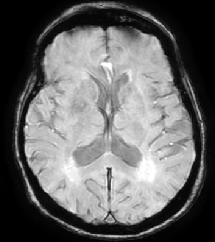} & \includegraphics[width=0.16\textwidth]{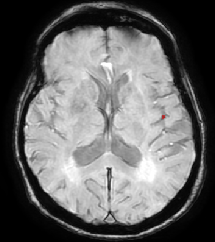}
 \\
  & \includegraphics[width=0.16\textwidth]{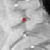} & \includegraphics[width=0.16\textwidth]{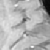} & \includegraphics[width=0.16\textwidth]{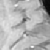} & \includegraphics[width=0.16\textwidth]{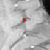}
 \\

  & \includegraphics[width=0.16\textwidth]{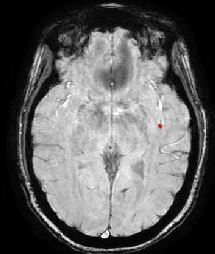} & \includegraphics[width=0.16\textwidth]{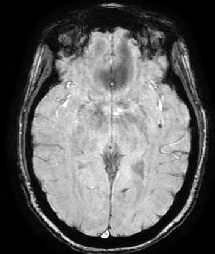} & \includegraphics[width=0.16\textwidth]{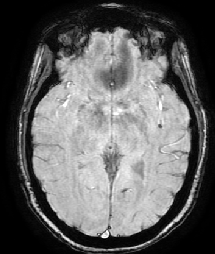} & \includegraphics[width=0.16\textwidth]{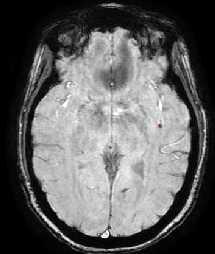}
 \\
  & \includegraphics[width=0.16\textwidth]{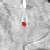} & \includegraphics[width=0.16\textwidth]{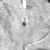} & \includegraphics[width=0.16\textwidth]{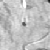} & \includegraphics[width=0.16\textwidth]{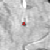}
 \\
  & \includegraphics[width=0.16\textwidth]{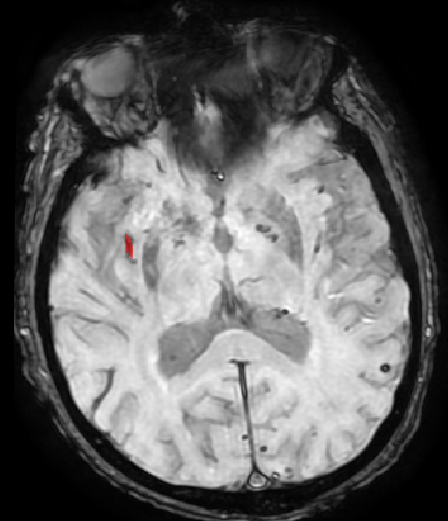} & \includegraphics[width=0.16\textwidth]{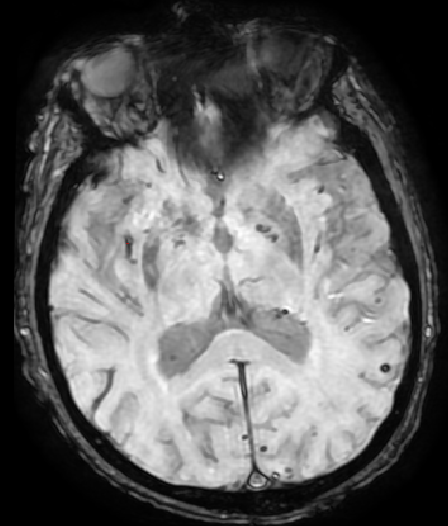} & \includegraphics[width=0.16\textwidth]{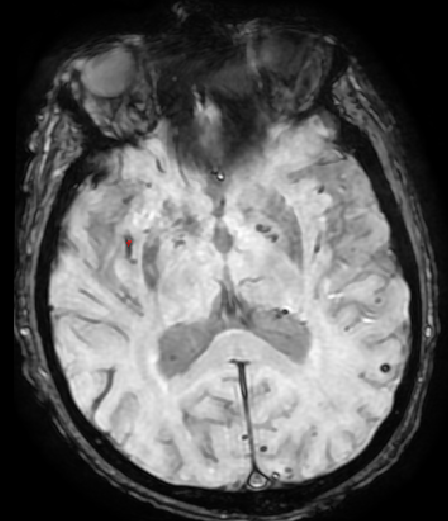} & \includegraphics[width=0.16\textwidth]{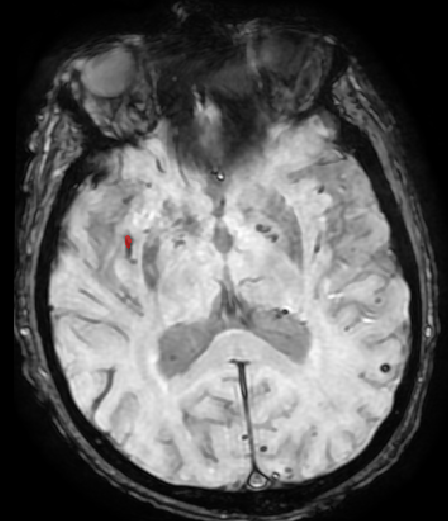}
 \\
   & \includegraphics[width=0.16\textwidth]{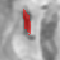} & \includegraphics[width=0.16\textwidth]{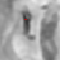}
  & \includegraphics[width=0.16\textwidth]{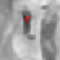} & \includegraphics[width=0.16\textwidth]{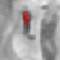}
 \\

\end{tabular}

\caption{SOTA prediction comparison. CLSTM, LLSTM and AttLLSTM predictions are compared to the groundtruth.  \label{fig:sotath}}
\end{figure*}

%  &&& & &&  & &&&& && && && & & (0.76) && &\\
%UpAttLLSTM  &MATAR2  && &\checkmark &&  $\mid$&1.0&&$\mid$&120.9& &$\mid$&0.2 &&$\mid$ &9.3& &$\mid$ &0.52 && $\mid$&0.86 \\
 % &&& & &&  & &&&& && &&&& &  &(0.60) && &\\

%\bottomrule
%\end{tabular}}
%\caption{Comparison of models with post-processing techniques and %Upsampling block. The highlight numbers are the best ones.  }
%\label{tab:resth3}
%\end{table*}

nnUnet struggles to segment the thrombi: less than 80$\%$ of the patients are detected. LLSTM outperforms CLSTM in the Dice score, even though the detection rate is lower. Adding to both recurrent methods, the attention module increases the performance. AttLLSTM has the highest detection rate (having 91$\%$ in the smaller cases) and Dice. Without using the distal occlusions, AttLLSTM misses 40$\%$ of the patients, and including them in the training improves all metrics. Also, the use of LLSTM and the attention block allows yo have an architecture with around 2 millions parameters, being smaller than the use of the equivalent using  CLSTM  (3 millions) and much smaller than nnUnet (30 millions).  We can see in Fig. \ref{fig:sotath} a visual comparison between the SOTA thrombi segmentations and groundtruths. We show SWAN with the masks superposed as the zoomed versions of the images, comparing CLSTM, LLSTM and AttLLSTM segmentations. We can see that the attention module helps to improve either the detection rate (first two examples) or the Dice score (last case), showing the importance and the added value of using the lesion informative modality. 

For the post-processing techniques, adding the lesion (Lesion) reduces the number of false positives in more than half of them but also reduces the detection rate for distal cases (some wrong close objects are chosen) for AttCLSTM. Finally, the threshold technique (Thr) allows to have an average Dice of 0.43 or higher for all. AttLLSTM gets the best performance (0.61 as Dice), detecting almost all the patients (missing less than 10$\%$ of distal thrombi). Indeed, the model is already better without the post-processing comparing it with nnUnet/AttCLSTM +lesion+thr in all metrics. 

\begin{table*}[!htb]
\centering
 { \scalebox{0.85}{\begin{tabular}{lccccc@{\hspace{-0.1ex}}c@{\hspace{-0.1ex}}cc@{\hspace{-0.1ex}}c@{\hspace{-0.1ex}}cc@{\hspace{-0.1ex}}c@{\hspace{-0.1ex}}cc@{\hspace{-0.1ex}}c@{\hspace{-0.1ex}}cc@{\hspace{-0.1ex}}c@{\hspace{-0.1ex}}cc@{\hspace{-0.1ex}}c@{\hspace{-0.1ex}}c}
\toprule
Grad.  &  Pretrained& Test dataset &Postprocessing & PHASE & \multicolumn{6}{c}{False Positives}& \multicolumn{6}{c}{False Negatives} &\multicolumn{3}{c}{Dice}&\multicolumn{3}{c}{Det. } \\ 
mod. \hspace{0.1in}&&&Big&&  \multicolumn{3}{c}{Count} & \multicolumn{3}{c}{Size}  & \multicolumn{3}{c}{Count} &  \multicolumn{3}{c}{Size} & & \\
drop.& &&&& P&$\mid$&  D& P  & $\mid$ &   D & P  &$\mid$& D  &P  &$\mid$ &  D  & P &$\mid$& D  & P &$\mid$& D 
\\
\midrule

& &CHSF+MATAR & & \fullcirc&2.1&  $\mid$& 2.2&34.2&$\mid$&56.9&0.1 &$\mid$& 0.2&0.1&$\mid$ &62.4& 0.62&$\mid$ & 0.58 &1.00& $\mid$&0.93\\
&&CHSF+MATAR &\checkmark & \fullcirc &\textbf{0.3}&  $\mid$& \textbf{0.1}&\textbf{7.8}&$\mid$&\textbf{12.5}&\textbf{0.1} &$\mid$& \textbf{0.3}&\textbf{0.1}&$\mid$ &\textbf{65.4}& \textbf{0.65}&$\mid$ &\textbf{ 0.65} &\textbf{1.00}& $\mid$&\textbf{0.86}\\
&  &CHSF+MATAR&  &  \emptycirc &0.6&$\mid$&0.6 & 14.4&$\mid$&18.5 &0.4 &$\mid$&0.6 &135.8 &$\mid$&172.3 & 0.24&$\mid$&0.29&0.8&$\mid$& 0.57\\ 

&&CHSF+MATAR & && 3.4& $\mid$&3.8&111.9&$\mid$&98.6& 0.13&$\mid$& 0.25&0.13&$\mid$ &75.8& 0.48&$\mid$ &0.39 &1.00& $\mid$ & 0.92\\

  &&FOCH (SWAN)& & &5.8 & $\mid$& &158.3 &$\mid$& &0.86  &$\mid$& & 198.5&$\mid$ & &  0.21&$\mid$ & &0.71 & $\mid$ &  \\

 &&FOCH (SWI)& &&  5.0& $\mid$&&99.7  &$\mid$& &  0.46&$\mid$& &187.3&$\mid$ && 0.33&$\mid$ & &0.88& $\mid$ &  \\

 \checkmark& &CHSF+MATAR  && \fullcirc &3.1&  $\mid$& 1.6&54.1&$\mid$&58.8& 0.2 &$\mid$& 0.3&40.8&$\mid$ &65.4& 0.47&$\mid$ & 0.58 & 0.90& $\mid$&0.86\\
 
 \checkmark& & CHSF+MATAR& & \emptycirc& 3.7 & $\mid$&3.0&62.5&$\mid$&132.6&0.2 &$\mid$& 0.3&40.8&$\mid$ &65.4& 0.44&$\mid$ & 0.42 &0.90& $\mid$&0.86\\ 
 
 \checkmark  & \checkmark  &CHSF+MATAR&  & \fullcirc& 3.2&  $\mid$& 3.4&76.9&$\mid$&86.7& 0.1 &$\mid$& 0.2&0.1&$\mid$ &62.4& 0.54&$\mid$ & 0.49 & 1.00& $\mid$&0.92\\
\checkmark  & \checkmark  &CHSF+MATAR&  \checkmark  & \fullcirc &0.4&  $\mid$& 0.6&29.4&$\mid$&83.8&  0.2&$\mid$& 0.4&37.4&$\mid$ &74.9& 0.60 &$\mid$ & 0.57 & 1.00& $\mid$&0.79\\
  
 \checkmark & \checkmark   &CHSF+MATAR&  & \emptycirc &5.0&  $\mid$&4.7 &94.7&$\mid$&101.6&  0.1&$\mid$&0.29 &0.1&$\mid$ &65.4&0.45 &$\mid$ &0.41  &1.00& $\mid$&0.86\\

 & &FOCH (SWAN)& &\emptycirc  &0.8& $\mid$& &23.6 &$\mid$& &1.0  &$\mid$& & 334.5&$\mid$ & &  0.14&$\mid$ & &0.45 & $\mid$ &  \\
&  &FOCH (SWI)& &\emptycirc  &1.1& $\mid$&&33.0  &$\mid$& &  1.1&$\mid$& &1007.1&$\mid$ && 0.04&$\mid$ & &0.32& $\mid$ &  \\ 
 \checkmark &&FOCH (SWAN)& &\emptycirc  &3.6 & $\mid$& &62.2 &$\mid$& &0.64  &$\mid$& & 90.5&$\mid$ & &  0.32&$\mid$ & &0.82 & $\mid$ &   \\ 
 \checkmark& &FOCH (SWI)& &\emptycirc   &4.1& $\mid$& &75.6 &$\mid$& &0.58  &$\mid$& & 311.6 &$\mid$ & &  0.21&$\mid$ & &0.77 & $\mid$ &  \\ 
  \checkmark& \checkmark   &FOCH (SWAN)& &\emptycirc& 4.2 & $\mid$& &101.6 &$\mid$& &0.55  &$\mid$& & 81.5&$\mid$ & &  0.37&$\mid$ & &0.90 & $\mid$ &   \\ 
   \checkmark& \checkmark   &FOCH (SWAN) &  \checkmark&\emptycirc &\textbf{1.1} & $\mid$& &\textbf{139.4} &$\mid$& &\textbf{0.64}  &$\mid$& & \textbf{174.2}&$\mid$ & &  \textbf{0.42} &$\mid$ & &\textbf{0.82} & $\mid$ &   \\ 
   
  \checkmark& \checkmark   &FOCH (SWI)& &\emptycirc  & 3.9& $\mid$& &71.5 &$\mid$& &0.55  &$\mid$& & 284.1 &$\mid$ & &  0.28&$\mid$ & &0.80 & $\mid$ & \\

\checkmark& \checkmark   &FOCH (SWI)  & \checkmark&\emptycirc &\textbf{0.8}& $\mid$& &\textbf{77.4} &$\mid$& &\textbf{0.58}  &$\mid$& & \textbf{311.6} &$\mid$ & &  \textbf{0.29} &$\mid$ & &\textbf{0.77}  & $\mid$ & \\

\bottomrule
\end{tabular}}}
{\caption{UpATTLLSTM thrombi segmentation results using SWI and SWAN. In FOCH, the total number of SWAN patients is 12 and of SWI is 31. $\fullcirc$ means that the modality is used in the test and $\emptycirc$ means that it is replaced by a black image.  \label{tab:missth1}}}
\end{table*}

The inclusion of MATAR2 dataset, as it adds more distal thrombi cases in the training, improves the performance in the distal cases (Dice = 0.61) but it reduces the performance on proximal cases, even reducing the detection rate. However, the Upsampling block helps to improve both dice, arriving at an average Dice score of 0.60, which is better than AttLLSTM with CHSF+MATAR and CHSF+MATAR+MATAR2 and is close to AttLLSTM with the two post-processing techniques. The results show the positive impact of using the upsampling block, as the Dice augments without the need of any post-processing technique.

We can see that choosing the biggest object improves the results: the number of false positives is indeed reduced.  The Dice score arrives at 0.65 in either distal or proximal thrombus. To conclude, we have proposed an architecture, UpAttLLSTM, that produces the best results when they are compared to several SOTA methods. The attention block allows for improvement in the detection rate, as the modalities are merged properly, the recurrent network allows to learn the spatial dependencies of the slices, and finally, the upsampling block improves the Dice score. However, only a monocentric dataset has been used for training and evaluation, so the machine's robustness is still not achieved.  %When we evaluate on MATAR2, where a greater variety of thrombi and then a more realistic scenario is tested, the Dice arrives at 0.52 on distal cases. As in that dataset, several clinical data is available, we check the performances obtained per thrombus location:
%\begin{itemize}

%\item Distal  (MCA) (M2d, M2p, M3/M4): 86 $\%$ detection rate and dice = 0.52

 %\item ACA (A1,A2) : 0 $\%$   detection rate and dice = 0.0

%\item PCA (P1,P2): 60 $\%$  detection rate and dice = 0.21
%\end{itemize}

%The performance under less common thrombus (ACA or PCA) is considerably decreased, detecting 0\% of the thrombus considered as ACA. Indeed, during training, they represent approximately 1\% of the cases, which explains the low detection rate achieved. 

 %Even though it lacks of generalisation for rare thrombus, as ACA or PCA, we have shown that it is consistent with the modalities' relationships: when the treatment has a big effect, the model faces problems in finding the thrombus. 
\subsection{Multicentric thrombi segmentation}
 \label{sec:missth}

We incorporate gradual modality dropout for the thrombus segmentation. In that case, UpAttLLSTM robustness to  PHASE missing is tested using FOCH dataset, where that modality is absent.  FOCH, being a multicentric dataset, has PHASE absent and in some cases the modality \ac{swi} replaces the modality \ac{swan}. Due to its small annotated size (43 images), we do not use it for the training, leaving it aside just for testing. Using $p$=0.5, two approaches are tested: starting the training from scratch, allowing $g(t)$ to start from the beginning and a transfer learning approach, where the training starts with the best UpAttLLSTM model obtained, using it as a pretrained architecture. The second approach reduces the training time, and only 400 epochs are added to the training to manage to include the gradual function. 

The results of these two approaches under missing and non-missing scenarios on CHSF, MATAR and FOCH, with either \ac{swi} or \ac{swan} are shown in Table \ref{tab:missth1}.

Without using any method, the performance decreases when PHASE is missing in CHSF and MATAR, having a Dice lower than 0.30 and almost no thrombi are detected on FOCH, where the detection rate is lower than 40\%. Also, in the models where PHASE is retired from the architecture (row 3-5), all performances are reduced. However, gradual modality dropout improves considerably  the detection rate and Dice score.

Comparing the approach of transfer learning or training from scratch, the best model is obtained thanks to transfer learning, where the detection rate on FOCH with SWAN images is 0.90\% and with SWI is 80\%, both dice scores are around 0.30. With this technique, in CHSF and MATAR, the dice are reduced to 0.52, while without the technique was around 0.60.

As done in the thrombi segmentation section, we include the post-processing technique of choosing the biggest predicted object to try to reduce the number of false positives for the transfer learning approach. Even though the model's performance decrease slightly due to the gradual modality dropout (Dice of 0.65 vs 0.57), it allows to have a good detection rate over the unseen dataset FOCH. It detects around 80\% of the thrombus with a Dice of 0.42 when SWAN is present and of 0.29 when SWI replaces SWAN, both having few false positives. 
\begin{figure*}[!htb]
\centering
\begin{tabular}{c@{\hspace{0.2in}}c@{\hspace{0.2in}}c@{\hspace{0.2in}}c@{ }c@{}c@{ }}

 &Groundtruth & UpAttLLSTM & UpAttLLSTM  \\
 & & & +mod drop \\
 & \includegraphics[width=0.17\textwidth]{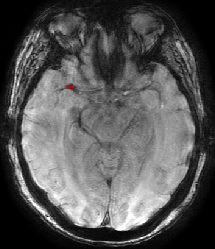} & \includegraphics[width=0.17\textwidth]{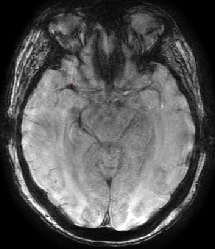} & \includegraphics[width=0.17\textwidth]{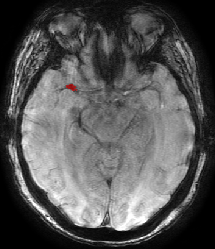} 
 \\
  & \includegraphics[width=0.17\textwidth]{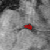} & \includegraphics[width=0.17\textwidth]{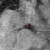} & \includegraphics[width=0.17\textwidth]{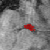} 
 \\
  & \includegraphics[width=0.17\textwidth]{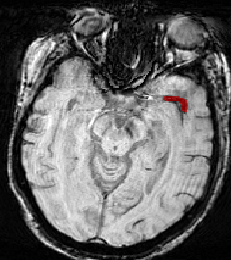} & \includegraphics[width=0.17\textwidth]{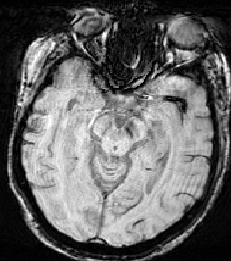} & \includegraphics[width=0.17\textwidth]{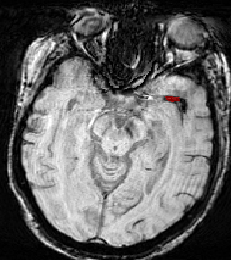} 
 \\
 
  & \includegraphics[width=0.17\textwidth]{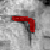} & \includegraphics[width=0.17\textwidth]{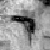}
  & \includegraphics[width=0.17\textwidth]{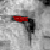} 
 \\
  & \includegraphics[width=0.17\textwidth]{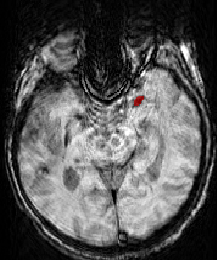} & \includegraphics[width=0.17\textwidth]{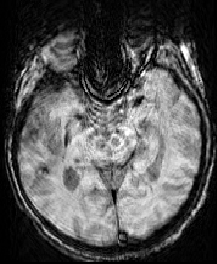} & \includegraphics[width=0.17\textwidth]{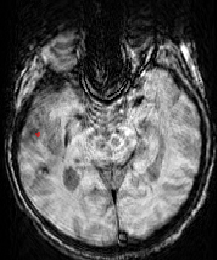} 
 \\
  & \includegraphics[width=0.17\textwidth]{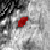} & \includegraphics[width=0.17\textwidth]{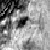} & \includegraphics[width=0.17\textwidth]{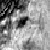} 
 \\
 
\end{tabular}

\caption{Segmentation examples. Comparison between groundtruth and the predictions obtained using  UpAttLLSTM and UpAttLLSTM + gradual modality dropout. \label{fig:sotath2}}
\end{figure*}
 In Fig. \ref{fig:sotath2} we can see some segmentation examples, showing the groundtruth, the segmentation obtained using the proposed architecture and incorporating also gradual modality dropout. For visualization reasons, the zoomed thrombus is included. Either using \ac{swi} or \ac{swan}, gradual modality dropout allows to improve the detection rate and Dice score (first two patients) as some thrombus are only detected by the model when the technique is incorporated. However, there are still some difficult thrombus that are still undetectable, as shows the last patient example.

The proposed method allows for improving the results when absent data and new centers are present in the evaluation, which is common in clinical practice for either the lesion or thrombi segmentation. With transfer learning, in addition to getting better results than training from scratch, the total training time is reduced. This method can be applied to any architecture, being robust and easy to adapt to any training strategy.

\section{Conclusion}

We have proposed a model that merges the lesion and thrombi informative MRI modalities through an attention-based architecture. This fusion technique, followed by a light recurrent network, obtains better results than the comparable SOTA methods, obtaining an average Dice of 0.65. As the multicentric robustness for the thrombus segmentation arises a missing modality, due to the differences in the MRI machine's brands, we propose gradual modality dropout. This proposed technique is included during training, without any adaptation to the architecture and allows to increase the performance when one modality is missing, without having a big impact in the case where all information is present. Thanks to it, our architecture is able to predict almost 80\% of the thrombus when PHASE is missing and the detection rate is reduced only by 5\% when SWAN is replaced by SWI.

 In conclusion, UpAttLLSTM with gradual modality dropout  produces promising results in the hyper-acute stroke phase, segmenting distal and proximal clots under a multicentric scenario. This detection could save time as its prediction is in less than 3 minutes, help to choose a treatment, providing the shape and volume information, which could help to  estimate the survival of the patients. 

\section*{Acknowledgements}
This work was granted access to the HPC resources of IDRIS under the allocation 2026-AD011014372R2 made by GENCI. »

%\appendix
%%\section{Example Appendix Section}
%\label{app1}


%%       \citet{<label>} ==> Lamport [21]
%%       \citep{<label>} ==> [21]
%%
%Example citation, See \citet{lamport94}.

  \bibliographystyle{elsarticle-num-names} 
\bibliography{bib}



\end{document}